\documentclass{article}

\PassOptionsToPackage{numbers, compress}{natbib}
    \usepackage[preprint]{neurips_2025}

\usepackage[utf8]{inputenc} % allow utf-8 input
\usepackage[T1]{fontenc}    % use 8-bit T1 fonts
\usepackage{hyperref}       % hyperlinks
\usepackage{url}            % simple URL typesetting
\usepackage{booktabs}       % professional-quality tables
\usepackage{amsfonts}       % blackboard math symbols
\usepackage{nicefrac}       % compact symbols for 1/2, etc.
\usepackage{microtype}      % microtypography
\usepackage{xcolor}         % colors
\usepackage{amsmath}
\usepackage{mathtools} % amsmath with fixes and additions
\usepackage{booktabs} % commands to create good-looking tables
\usepackage{caption}
\usepackage{tikz} % nice language for creating drawings and diagrams
\usetikzlibrary{arrows.meta}

\newcommand{\ve}{\boldsymbol{e}}

\newcommand{\vtheta}{\boldsymbol{\theta}}
\newcommand{\vphi}{\boldsymbol{\phi}}
\newcommand{\vz}{\boldsymbol{z}}
\newcommand{\vx}{\boldsymbol{x}}
\newcommand{\vmu}{\boldsymbol{\mu}}
\newcommand{\cL}{\mathcal{L}}
\newcommand{\E}{\mathbb{E}}
\newcommand{\KL}{D_{\mathrm{KL}}}

\title{CLEAR: Unlearning Spurious Style-Content Associations with Contrastive LEarning with Anti-contrastive Regularization}

\author{%
  Minghui Sun \\
  Department of Biostatistics \& Bioinformatics\\
  Duke University\\
  Durham, NC 27705 \\
  \texttt{minghui.sun@duke.edu} \\
  \And
  Benjamin Goldstein \\
  Department of Biostatistics \& Bioinformatics \\
  Duke University\\
  Durham, NC 27705 \\
  \texttt{ben.goldstein@duke.edu} \\
  \AND
  Matthew Engelhard \\
  Department of Biostatistics \& Bioinformatics \\
  Duke University\\
  Durham, NC 27705 \\
  \texttt{m.engelhard@duke.edu} \\
}

\begin{document}

\maketitle

\begin{abstract}
  Learning representations unaffected by superficial characteristics is important to ensure that shifts in these characteristics at test time do not compromise downstream prediction performance. For instance, in healthcare applications, we might like to learn features that contain information about pathology yet are unaffected by race, sex, and other sources of physiologic variability, thereby ensuring predictions are equitable and generalizable across all demographics. Here we propose \textbf{C}ontrastive \textbf{LE}arning with \textbf{A}nti-contrastive \textbf{R}egularization (CLEAR), an intuitive and easy-to-implement framework that effectively separates essential (\textit{i.e.}, task-relevant) characteristics from superficial (\textit{i.e.}, task-irrelevant) characteristics during training, leading to better performance when superficial characteristics shift at test time. We begin by supposing that data representations can be semantically separated into task-relevant \textit{content} features, which contain information relevant to downstream tasks, and task-irrelevant \textit{style} features, which encompass superficial attributes that are irrelevant to these tasks, yet may degrade performance due to associations with \textit{content} present in training data that do not generalize. We then prove that our anti-contrastive penalty, which we call Pair-Switching (PS), minimizes the Mutual Information between the \textit{style} attributes and \textit{content} labels. Finally, we instantiate CLEAR in the latent space of a Variational Auto-Encoder (VAE), then perform experiments to quantitatively and qualitatively evaluate the resulting CLEAR-VAE over several image datasets. Our results show that CLEAR-VAE allows us to: (a) swap and interpolate \textit{content} and \textit{style} between any pair of samples, and (b) improve downstream classification performance in the presence of previously unseen combinations of \textit{content} and \textit{style}. Our code is at: \url{https://github.com/scotsun/clear-vae}.
\end{abstract}

\section{Introduction}\label{sec:intro}

Information present in a given feature vector can be semantically separated into two components: \textit{content} and \textit{style} \citep{mathieu2016disentangling, bouchacourt2018mlvae, hamaguchi2019rare, abid2019contrastive, louiset2023sepvae}. When ground truth outcome labels are provided, we may define \textit{content} as the information directly related to the outcome label, whereas \textit{style} reflects additional variability that is not relevant to the outcome of interest, which may or may not correspond to a second set of labels in the dataset. 
For instance, in healthcare applications, the \textit{content} might contain clinical factors directly related to diagnoses or health outcomes, whereas the \textit{style} includes sources of physiologic variability, including those related to race and gender, that would not be associated with outcomes in a representative, unbiased sample. 
However, while unrelated to the labels in principle, the \textit{style} can nevertheless influence a model's classification performance and generalizability due to spurious correlations between content and style observed in training data, for instance due to health inequities. Such associations can be avoided by collecting an arbitrarily large, representative, unbiased dataset, but in practice they are common due to (1) the rarity of a given outcome of interest, (2) heterogeneous \textit{style} distributions across (biased) outcome groups, and (3) distribution shifts taking place between training and test time.

Several unsupervised representation learning methods have been developed in recent years to disentangle \textit{content} from \textit{style} \citep{higgins2017beta, burgess2018understanding, kim2018factorvae, chen2018isolating}. 
When effective, such methods allow downstream predictions to be unaffected by \textit{style}, in turn preventing spurious content-style associations from influencing prediction performance. 
These unsupervised methods achieve disentanglement by encouraging fully factorized encoded representations. However, if we are interested in using encoded information for downstream classification tasks, a key limitation of these methods is that the model still cannot directly identify the \textit{content} information from the encoded latent features. Moreover, recent work has shown that the disentanglement provided by these unsupervised approaches does not link to more efficient learning in downstream tasks \citep{locatello2019challenging, locatello2020commentary}. Taking the opposite extreme, disentanglement can be directly achieved in a doubly supervised setting, in which labels for both \textit{content} and \textit{style} are provided to the model. However, this strong degree of supervision is typically unavailable in a real-world scenario, because in practice, we are unlikely to have a complete set of \textit{style} labels that explicitly delineate unwanted sources of variability.

\begin{figure}
    \centering
    \includegraphics[width=0.68\linewidth]{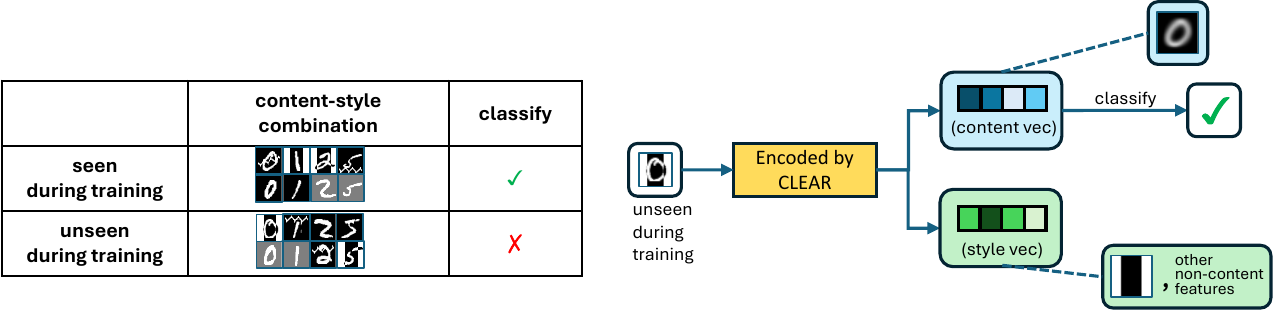}
    \caption{CLEAR-based disentanglement aids classification of unseen style-content combinations}
    \label{fig:fig1}
\end{figure}

In this work, therefore, we aim to effectively disentangle content and style in the much more common setting in which \textit{content} labels are available, but \textit{style} labels are not. We design a weakly supervised method that simultaneously learns \textit{content} representations -- guided solely by the outcome label -- and disentangles \textit{style} from \textit{content} without requiring \textit{style} annotations. Our method, which we call Contrastive LEarning with Anti-contrastive Regularization (CLEAR), has two primary components:

\textbf{First, use \textit{contrastive learning} to maximize MI between content features and outcome labels.} First, we use supervised contrastive framework to guide the representation learning of \textit{content} features. Based on the premise that each feature vector is composed of \textit{content} and \textit{style} semantics, we show that maximizing the mutual information (MI) between the content labels and content-specific latent factors naturally arises from ELBO maximization. Leveraging the established link between MI and contrastive learning \citep{oord2018representation, khosla2020supervised},
we use only a single set of labels, the so-called \textit{content} labels, to form contrastive pairs involved in the training objective.

\textbf{Second, use a novel \textit{anti-contrastive regularization} term to minimize MI between style features and outcome labels.} To effectively decouple \textit{content} and \textit{style} semantics in the absence of \textit{style} labels, we include an additional anti-contrastive regularization term, which we call Pair-Switching (PS), to minimize the MI between style-specific latent factors and content labels. We name it PS, since, to apply it, we need only flip the roles of positive and negative pairs in the original contrastive loss. Indeed, our approach was motivated by circumstances where (a) style distributions differ between labels, yet (b) it is reasonable to suppose that style distributions would be consistent between labels in a large, representative, unbiased dataset. For example, many medical diagnoses are associated with demographic characteristics in observational datasets due to disparities in access to care, but it is reasonable to suppose that in an ideal, unbiased dataset, such associations would not exist.

After defining CLEAR and establishing properties related to MI minimization, we instantiate CLEAR in the latent space of a Variational Auto-Encoder (VAE), resulting in CLEAR-VAE. While in principle, CLEAR can be paired with other generative models \textit{(e.g.}, GANs, diffusion models), our CLEAR-VAE implementation is advantageous due to (a) our focus on downstream prediction rather than image generation, and (b) the ease of VAE implementation and training, which allow us to distinguish the benefits of CLEAR from variability in performance associated with the choice of generative model.

We then conduct extensive experiments on multiple datasets to evaluate CLEAR-VAE’s disentanglement performance and its effectiveness in enhancing model generalizability. To begin with, in our qualitative analysis, we perform interpolation and swapping experiments originally proposed by \citet{mathieu2016disentangling}. We demonstrate that our approach achieves effective disentanglement, successfully extracting \textit{style} characteristics which are known to us \textbf{but unknown to the model} through a weakly supervised contrastive approach requiring only the \textit{content} labels. Finally, we quantitatively show that CLEAR-VAE can improve performance on downstream tasks in the presence of
%a downstream model's generalizability on 
out-of-distribution (OOD) samples with unseen combinations of \textit{content} and \textit{style}.

In summary, our contributions are as follows: 
\textbf{1)} We introduce a framework for evaluating benefits of disentanglement when style-content associations change at test time. \textbf{2)} We propose CLEAR, a novel, easy-to-implement, contrastive-learning-based framework for unlearning spurious style-content associations and therefore achieving disentanglement between the two semantics. \textbf{3)} We establish theoretical justification for CLEAR, analyze its PS penalty in comparison to existing mutual information (MI) minimization methods, and introduce several CLEAR variants that incorporate these MI minimization techniques. \textbf{4)} We perform experiments across five different datasets to demonstrate CLEAR can (a) swap and interpolate \textit{content} and \textit{style} between any pair of samples, and (b) improve downstream tasks' generalizability to OOD samples with unseen combinations of \textit{content} and \textit{style}.

\section{Related work}
\textbf{Semantic Disentanglement in VAE.} There are various frameworks to achieve semantic disentanglement in a VAE's latent representation. \citet{mathieu2016disentangling} achieve disentanglement through adversarial training on the VAE decoder. \citet{bouchacourt2018mlvae} and \citet{hosoya2018gvae} organize samples into groups based on ground truth labels within mini-batches and extract group-level content representations. \citet{hamaguchi2019rare} learn the \textit{content} through similarity regularization. \citet{abid2019contrastive, louiset2023sepvae} propose methods to distinguish target-specific salient (\textit{content}) semantics from common background (\textit{style}) semantics by utilizing contrastive pairs. These pairs consist of target samples and their corresponding background samples. The aforementioned methods are designed to separate latent representations into two parts: \textit{content} and \textit{style}. They provide evidence that an effective disentanglement can enhance model generalizability and therefore improve downstream tasks' performance on OOD combinations of \textit{content} and \textit{style} \citep{mathieu2016disentangling, bouchacourt2018mlvae, hosoya2018gvae, hamaguchi2019rare, louiset2023sepvae}. Different from \citet{mathieu2016disentangling, hamaguchi2019rare, abid2019contrastive, louiset2023sepvae}, in our framework, the VAE architecture does not require parallel branches of networks, nor do we need pre-paired data for model training. \citet{bouchacourt2018mlvae} and \citet{hosoya2018gvae} focus on regularizing \textit{content}, but we achieve semantic disentanglement by considering both \textit{content} and \textit{style}.

\textbf{Contrastive Learning.} Contrastive learning was originally developed as a form of supervised learning \citep{chopra2005maxmargin, koch2015siamese}. More recent methods based on InfoNCE have gained significant popularity in self-supervised learning (SSL) \citep{oord2018representation, he2020momentum, chen2020simple, damrich2022t}. In these methods, positive pairs are constructed by applying different random augmentations to the same samples, whereas negative pairs are constructed by applying random augmentations to different samples. These methods have been shown to maximize the MI between different representations corresponding to the different augmentations of the same data. In a fully supervised setting, however, pairs can be alternatively defined based on class labels, with positive pairs indicating that both samples belong to the same class, and negative pairs indicating that the samples come from different classes \citep{frosst2019analyzing, khosla2020supervised}. In general, a contrastive learning method aims to calibrate the similarity/distance between paired elements in an embedding space, encouraging positive pairs to be clustered together while negative pairs to remain far apart.

\textbf{Mutual Information Minimization.} Various numerical methods can be used to calculate a differentiable upper bound or estimate for the MI between content and style factors. Some approaches involve deriving a variational upper bound for the MI between two arbitrary variables, $I(X;Y)$, and estimating it through a separate parameterized neural network $p_{\psi}(x|y)$ \citep{poole2019l1out, cheng2020club}. Other methods reformulate mutual information as the KL divergence between the joint distribution and the product of the marginals -- known as the Total Correlation (TC) loss -- and approximate it using the density-ratio trick through a separate discriminator \citep{sugiyama2012density, kim2018factorvae, abid2019contrastive, louiset2023sepvae}. In the aforementioned methods, the auxiliary network and the main VAE model undergo alternating adversarial training. Appx. \ref{appx:mim} gives more details on the implementation of these methods. These techniques are formulated to minimize MI between latent semantics but also minimize the MI between style semantics and content labels through the data processing inequality \citep{cover1999elements, kirsch2024bridgingthedata}. Instead, the PS approach directly minimizes MI between style semantics and content labels. Additionally, the training complexity in PS is much smaller, as it does not rely on additional networks for adversarial learning.

\section{Method}\label{sec:method}
\subsection{CLEAR-VAE overview}

A vanilla VAE consists of two components: an encoder $q_{\phi}(\vz|\vx)$ that probabilistically maps an input $\vx$ to $\mathcal{N} \big( \vz ; \vmu_\phi(\vx) , \Sigma_{\phi}(\vx) \big)$ and a decoder $p_{\theta}(\vx|\vz)$ that reconstructs $\vx$ based on a stochastic latent representation $\vz$ \citep{kingma2013auto}. A standard VAE can be optimized under Evidence Lower Bound (ELBO):
\begin{align}
    \label{elbo}
     \cL_{\text{VAE}} &= -\text{ELBO} \notag\\ 
     &= -\E_{q_{\phi}}\left[ \log p_{\theta}(\vx|\vz) \right] + \KL ( q_{\phi}(\vz|\vx) \Vert p(\vz) )
\end{align}
In Eqn \ref{elbo}, the first term can be interpreted as the reconstruction loss: mean-squared error if $\vx$ is assumed to follow a Gaussian distribution, or binary cross-entropy if $\vx$ is binary. Then second term regularizes the KL divergence between the encoder and $p(\vz)$, the prior of $\vz$, which is usually assumed to be $\mathcal{N}(0, \boldsymbol{I})$. The reparameterization trick is a key technique that separates the deterministic and stochastic parts of sampling from a latent distribution and allows back-propagation to bypass the randomness in the probabilistic model \citep{kingma2013auto}. $\beta$-VAE is a straightforward extension of the VAE designed to learn unsupervised disentangled representations from data by multiplying the KL regularization term with a penalizing coefficient $\beta$ (typically, $\beta > 1$) \citep{higgins2017beta}. A stronger penalization can lead to a more factorized latent representation.

\begin{figure}[!h]
  \centering
  \begin{minipage}{0.30\textwidth}
    \centering
    \resizebox{!}{0.15\textheight}{%
        \begin{tikzpicture}[
    node distance=3cm, % Set distance between nodes
    every node/.style={
        circle, 
        draw, 
        minimum size=0.8cm, 
        font=\footnotesize % Smaller font size
    },
    ->, >=Stealth
]

% Nodes
\node (y) {$y$};
\node (zc) [right of=y, yshift=1.5cm] {$\boldsymbol{z}^{(c)}$};
\node (zs) [right of=y, yshift=-1.5cm] {$\boldsymbol{z}^{(s)}$};
\node (x) [right of=y, xshift=3cm] {$\boldsymbol{x}$};

% Arrows
\draw[->, bend left=15] (y) to node[midway, above, draw=none, yshift=-0.4cm] {$p(\boldsymbol{z}^{(c)}|y)$} (zc);
\draw[->, bend left=15] (zc) to node[midway, above, draw=none, yshift=-0.4cm] {$p_{\boldsymbol{\theta}}(\boldsymbol{x}|\boldsymbol{z}^{(c)})$} (x);
\draw[->, bend right=15] (zs) to node[midway, below, draw=none, yshift=0.4cm] {$p_{\boldsymbol{\theta}}(\boldsymbol{x}|\boldsymbol{z}^{(s)})$} (x);

% Backward arrows
\draw[<-, dashed, bend right=15] (y) to node[midway, below, draw=none, yshift=0.4cm] {$f(y|\boldsymbol{z}^{(c)})$} (zc);
\draw[<-, dashed, bend right=15] (zc) to node[midway, below, draw=none, yshift=0.6cm] {$q_{\boldsymbol{\phi}}(\boldsymbol{z}^{(c)}|\boldsymbol{x})$} (x);
\draw[<-, dashed, bend left=15] (zs) to node[midway, above, draw=none, yshift=-0.6cm] {$q_{\boldsymbol{\phi}}(\boldsymbol{z}^{(s)}|\boldsymbol{x})$} (x);

\end{tikzpicture}
    }
    \captionsetup{font=small}
    \caption*{(a) graphical representation}
  \end{minipage}
  \hfill
  \begin{minipage}{0.60\textwidth}
    \centering
    \includegraphics[height=0.15\textheight]{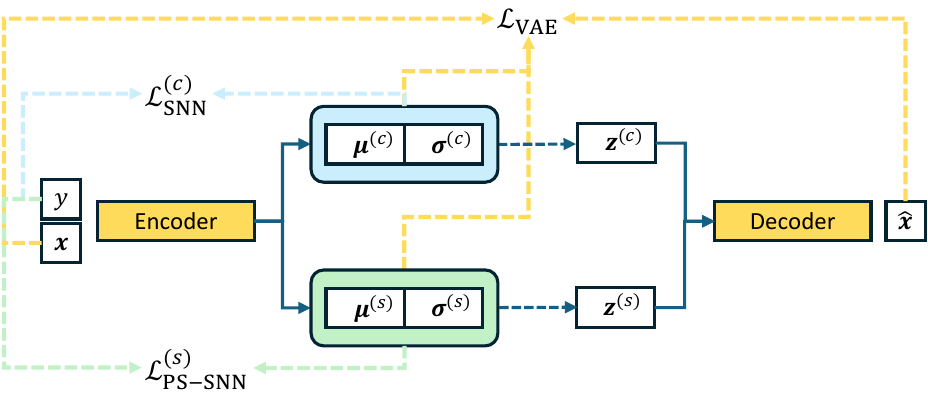}
    \captionsetup{font=small}
    \caption*{(b) schematic illustration}
  \end{minipage}
  \caption{Graphical representation and schematic illustration of CLEAR-VAE}
  \label{fig:clear-vae}
\end{figure}

The Markov graphical model in Fig.\ref{fig:clear-vae} (a) reflects our assumptions about the semantical disentanglement. The solid edges correspond to the generative distribution, while the dashed edges correspond to the posteriors' variational approximations. Input $\vx$ depend on $\vz^{(c)}$ and $\vz^{(s)}$. The content latent variables $\vz^{(c)}$ is intrinsically determined by $y$, whereas the style latent variables should be independent of the $y$. Following this graphical model, we can derive an ELBO as Eqn.~\ref{eq:elbo}:
\begin{align}
    \log p_{\vtheta}(x) &\ge \E_{\vz\sim q_{\vphi}}[\log p_{\vtheta}(\vx|\vz)] \notag \\
    &\quad - D_{\text{KL}}\Big(q_{\vphi}(\vz^{(c)}|\vx) \| p(\vz^{(c)})\Big)
    - D_{\text{KL}}\Big(q_{\vphi}(\vz^{(s)}|\vx) \| p(\vz^{(s)})\Big) \notag  \\
    &\quad + \E_{y\sim p,z^{(c)}\sim q_{\vphi}}\left[\log \frac{f(y|\vz^{(c)})}{p(y)}\right]
    \label{eq:elbo}
\end{align}
Appx~\ref{appx:elbo-decomp} provides more details about the derivation of the ELBO. The first three terms in the ELBO constitute the regular VAE loss function, namely, $\cL_{\text{VAE}}$. The last term is not $I(y;\vz^{(c)})$ but relates to the InfoNCE loss \citep{oord2018representation}, which in turn can be further reduced to a modified version of Soft Nearest Neighbor (SNN) loss \citep{frosst2019analyzing}. The original $L_2$ distance is replaced by cosine similarity. Appx. ~\ref{appx:snn-derivation} provides detailed discussion and derivation of the contrastive loss denoted by $\cL_{\text{SNN}}$. Similar to contrastive learning in the SSL \citep{wang2020understanding, chen2020simple}, supervised contrastive learning encourages the latent representations $\vz^{(c)}$ from the same class to cluster together while pushing those from different classes apart \citep{khosla2020supervised}. Minimizing this term encourages to maximize $I(y;\vz^{(c)})$. To achieve disentanglement between $\vz^{(c)}$ and $\vz^{(s)}$, we implement pair-switched SNN loss, $\mathcal{L}_{\text{PS-SNN}}$. Minimizing it encourages to minimize $I(y;\vz^{(s)})$, as shown in Sec.~\ref{ps}. Fig.~\ref{fig:clear-vae}~(b) illustrates the schematic workflow.

\begin{figure}
  \centering
  \begin{minipage}{0.3\textwidth}
    \centering
    \includegraphics[height=2.5cm]{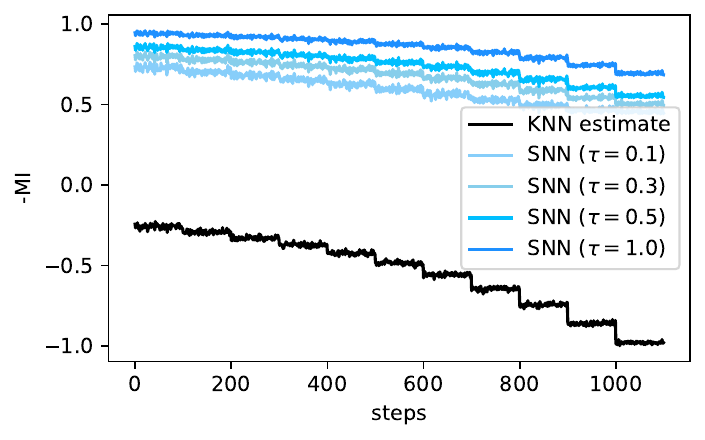}
    \vspace{-0.25cm}
    \captionsetup{font=tiny}
    \hspace{0.5cm}\caption*{\hspace{0.5cm}(a) SNN for MI maximization}
  \end{minipage}
  \begin{minipage}{0.3\textwidth}
    \centering
    \includegraphics[height=2.5cm]{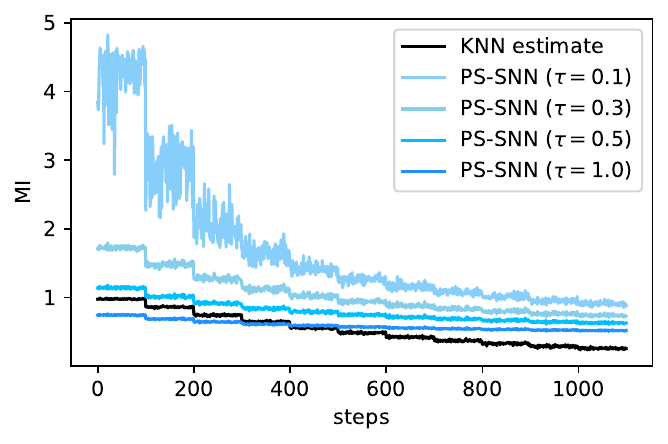}
    \vspace{-0.25cm}
    \captionsetup{font=tiny}
    \hspace{0.5cm}\caption*{\hspace{0.5cm}(b) PS-SNN for MI minimization}
  \end{minipage}
  \caption{MI maximization and minimization on simulated data from Gaussian distributions.}
  \label{fig:mi-simulation}
\end{figure}

Therefore, we optimize our CLEAR-VAE model using the following loss function.
\begin{equation}
    \label{clear-vae-loss}
    \cL = \cL_{\text{VAE}}(\beta) + \alpha_1 \cL_{\text{SNN}}^{(c)} + \alpha_2 \cL_{\text{PS-SNN}}^{(s)}
\end{equation}
To summarize, $\cL_{\text{VAE}}(\beta)$ is the $\beta$-VAE objective function with KL divergence terms for both content and style. $\cL_{\text{SNN}}^{(c)}$ is a contrastive term to maximize $I(y;\vz^{(c)})$. $\cL_{\text{PS-SNN}}^{(s)}$ is our anti-contrastive term to minimize $I(y;\vz^{(s)})$. In practice, we set $\alpha_1 = \alpha_2$ to simplify hyperparameter tuning.

We perform a simulation study to numerically verify the efficacy of $\cL_{\text{SNN}}$ and $\cL_{\text{PS-SNN}}$. Specifically, we sample $\vz$ from a Gaussian mixture, with $y$ indicating the component assignment. Since $I(y,\vz)$ does not have a closed-form expression, we approximate it using the KNN estimator \citep{ross2014mutual}. In Fig.~\ref{fig:mi-simulation} (a), we reduce the variances every 100 steps to increase $I(y, \vz)$; in (b), we increase them to decrease $I(y, \vz)$. Appx.~\ref{appx:mi-simulation} provides more detail.

\subsection{Multi-positive pair Contrastive Learning}

The InfoNCE-based loss have been commonly used as the contrastive objective in SSL, which allows only one positive pair per observation within a batch \citep{oord2018representation, he2020momentum, chen2020simple, damrich2022t}. In contrast, SNN and SupCon allows each sample to have multiple positive pairs within a batch \citep{frosst2019analyzing, khosla2020supervised}. In our $\cL_{\text{SNN}}$, we replace the $L_2$ distance with cosine similarity $\mathrm{sim}(\cdot,\cdot)$. This change is aligned with our derivation in Appx. \ref{appx:snn-derivation}; moreover, our experiments show that using cosine similarity leads to better disentanglement (Tab. \ref{tab:sims-metric}). Different from a SSL contrastive loss that maximizes the MI between different representations of the same data, a supervised contrastive loss maximizes the MI between data representations and supervised class labels. In Appx. \ref{appx:elbo-mi} and \ref{appx:snn-derivation}, we develop the connection between $\cL_{\text{SNN}}$ and MI.

As it has been shown in \citet{oord2018representation}, maximizing $\mathcal{L}_{\text{InfoNCE}}(y;\vz^{(c)})$ is equivalent to maximizing $\pi(c)$, the probability that sample $y$ is drawn from $f(y|\vz^{(c)})$ rather than $p(y)$, estimated as follows:
\begin{equation}
    \pi(c) = \frac{h(y, \vz^{(c)})}{\sum_{y'\in\mathcal{Y}}h(y', \vz^{(c)})}
    \label{eqn:prob-c}
\end{equation}
where $h(y,\vz^{(c)})$ is a learnable log-bilinear model to estimate the density ratio $\frac{f(y|\vz^{(c)})}{p(y)}$ \citep{oord2018representation}. Although $h(y, \vz^{(c)})$ is commonly replaced by cosine similarity in \citep{he2020momentum, chen2020simple, wang2020understanding}, we retain generality in the following derivation. In our notation, the InfoNCE loss on the content representation is:
\begin{equation}
    \cL_{\text{InfoNCE}}^{(c)} = -\E_{\vz^{(c)}\sim q_{\phi}, y\sim p} \left[ \log \frac{h(y, \vz^{(c)})}{\sum_{y'\in\mathcal{Y}}h(y', \vz^{(c)})} \right]
    \label{eqn:infonce}
\end{equation}
Then, we derive $\cL_{\text{SNN}}^{(c)}$ based on $\cL_{\text{InfoNCE}}^{(c)}$ and the learnable function $h(\cdot)$ is simplified to $\mathrm{sim}(\cdot,\cdot)$. Appx. \ref{appx:snn-derivation} provides more details. Our modified SNN loss is
\begin{equation}
\cL_{\text{SNN}}^{(c)} 
% &= \frac{1}{N} \sum_{i=1}^{N} -\log \frac{\sum_{y_j=y_i}h(y_j, \vz_i^{(c)})}{\sum_{y_j}h(y_j, \vz_i^{(c)})} \notag \\
= \frac{1}{N} \sum_{i=1}^{N} -\log \frac{\mathrm{pos}_i^{(c)}}{\mathrm{neg}_i^{(c)} + \mathrm{pos}_i^{(c)}}
\label{eqn:approx-snn}
\end{equation}
where
{
\begin{equation}
    \text{pos}_i^{(c)} \coloneqq \sum_{j\ne i} \mathbb{I}_{[y_i = y_j]}\exp \left\{ \mathrm{sim}(\vz^{(c)}_i,\vz^{(c)}_j)/\tau \right\},\quad
    \text{neg}_i^{(c)} \coloneqq \sum_{j\ne i} \mathbb{I}_{[y_i \ne y_j]}\exp \left\{ \mathrm{sim}(\vz^{(c)}_i,\vz^{(c)}_j)/\tau  \right\} 
\end{equation}
}

Apart from cosine similarity, we also can use distributional metrics on the entire encoded distributions $p_{\vtheta}(\vz_i^{(c)}|\vx_i)$ and $p_{\vtheta}(\vz_j^{(c)}|\vx)$ in $\cL_{\text{SNN}}^{(c)}$. \citet{hamaguchi2019rare} suggest to use metrics accounting for the latent variances to align positive pairs, such as Mahalanobis Distance \citep{mahalanobis2018generalized} or Jeffrey Divergence \citep{jeffreys1946invariant}. However, the Eqn. (\ref{eq:elbo}) favors Monte Carlo estimation with cosine similarity, and our empirical results in \ref{tab:sims-metric} shows variance-dependent distributional metrics gives reduced performance.

\subsection{Pair-Switching as Anti-Contrastive Regularization for MI Minimization}
\label{ps}

Along with using $\cL_{\text{SNN}}^{(c)}$ to learn meaningful $\vz^{(c)}$, we introduce an anti-contrastive regularization to encourage the style features ``unlearn'' the content label, thereby further encouraging disentanglement. In this subsection, we describe this anti-contrastive regularization term, which we call \textit{pair-switching} (PS), and prove why it is a numerically stable method to minimize the $I(y;\vz^{(s)})$.

Recall that the SNN loss of the $i$-th sample's \textit{content} representation can be abbreviated as 
\begin{equation}
    l_{i}^{(c)} = -\log \frac{\text{pos}_{i}^{(c)}}{\text{pos}_{i}^{(c)} + \text{neg}_{i}^{(c)}}
    \label{eqn:li}
\end{equation}
The loss is always positive. Since our goal is to dissociate $\vz^{(s)}$ with $y$, we might consider minimizing $-l_i^{(s)}$, the negative of the loss defined in equation Eqn. \ref{eqn:li}, with the content features $\vz^{(c)}$ replaced by the style features $\vz^{(s)}$. However, this introduces an unbounded negative value into the objective function, thereby complicating the minimization of $\cL_{\text{VAE}}$. Moreover, minimizing $-\cL_{\text{SNN}}^{(s)}$, the lower bound, does not guarantee that the $I(y;\vz^{(s)})$ is also minimized.

Therefore, we design PS to improve numerical stability while also providing a mathematically grounded approach to minimize MI. We proceed by following Eqn. \ref{eqn:prob-c}, but rather than maximizing $\pi(s)$ as in InfoNCE, which would imply that $y$ is more likely drawn from $f(y|\vz^{(s)})$ than $p(y)$, we instead want to maximize $1 - \pi(s)$, the probability that $y$ is drawn from $p(y)$ rather than $f(y|\vz^{(s)})$:
\begin{equation}
    1 - \pi(s) = 1 - \frac{h(y, \vz^{(s)})}{\sum_{y'\in\mathcal{Y}}h(y', \vz^{(s)})}
    \label{eqn:prob-s}
\end{equation}
Following \citet{oord2018representation}, maximizing Eqn. \ref{eqn:prob-s} is equivalent to minimizing the following pair-switched InfoNCE loss:
\begin{equation}
    \cL_{\text{PS-InfoNCE}}^{(s)} = -\E_{\vz^{(s)}\sim q_{\phi}, y\sim p} \left[ \log \frac{\sum_{y'\in \mathcal{Y}\setminus\{y\}}h(y', \vz^{(s)})}{\sum_{y'\in\mathcal{Y}}h(y', \vz^{(s)})} \right]
    \label{eqn:ps-infonce}
\end{equation}
Since $y$ is a categorical variable, many samples from $\mathcal{Y}$ can share the same value. Thus, to avoid contrasting $\vz^{(s)}$ with other $y'$ where $y'=y$, we remove all the pairs $(y',\vz^{(s)})$ for which $y'=y$ in the numerator. Then, we obtain the pair-switched SNN loss. Appx. \ref{ps-mi-ub} provides more details.
\begin{equation}
    \mathcal{L}_{\text{PS-SNN}}^{(s)}
    % &= \frac{1}{N} \sum_{i=1}^{N} - \log \frac{\sum_{y_j\ne y_i}h(y_j, \vz^{(s)}_i)}{\sum_{j}h(y_j, \vz^{(s)}_i)}  \notag \\
    = \frac{1}{N}\sum_{i=1}^{N} -\log\left(\frac{\mathrm{neg}_i^{(s)}}{\mathrm{pos}_i^{(s)} + \mathrm{neg}_i^{(s)}}\right)
\end{equation}
We call these the ``pair-switching" losses, as in practice they correspond to reversing the roles of positive and negative pairs in the original contrastive losses. The optimization will favor samples being selected from $p(y)$ rather than $f(y|\vz^{(s)})$, thereby minimizing $I(y;\vz^{(s)})$. Note that $\cL_{\text{PS-SNN}}^{(s)}$ is always non-negative, providing numeric stability. In Appx. \ref{ps-mi-ub}, we provide a complete proof that $\cL_{\text{PS-SNN}}^{(s)}$ is an upper bound of $I(y; \vz^{(s)}) - \log(N)$. Hence, minimizing $\cL_{\text{PS-SNN}}^{(s)}$ encourages the representation $\vz^{(s)}$ to contain minimal information about the $y$.

\section{Experiments}\label{sec:experiment}

\subsection{Datasets}
Tab. \ref{tab:dataset-def} provides the definitions of \textit{content} and \textit{style} in each dataset. We perform two sets of experiments. First, we include all combinations of content and style in both training and testing sets to qualitatively illustrate semantic disentanglement. Second, we evaluate CLEAR-VAE encoder's generalizability on OOD data with unseen \textit{styles}. We generate random train-test split of each dataset while ensuring that the \textit{styles} differ between training and testing sets for all classification tasks. We design a metric called Group Mutual Information Gap (gMIG) introduced in Appx. \ref{sec:gmig} to gauge the degree of the disentanglement. Appx.~\ref{appx:arch} lists model architectures for all datasets.

\textbf{Styled-MNIST \& Colored-MNIST.} We enhance the MNIST dataset \citep{lecun1998gradient} with more noticeable style features using the corruption transforming methods from \citet{mu2019mnist}. Each digit is randomly assigned to a transformation from \{identity/unchanged, stripe, zigzag, canny edge, tiny scaled, brightness\}. Similarly, we also enhance the MNIST by varying the color of digits. We randomly assign each digit to a color from \{red, green, blue, yellow, cyan, magenta\}.

\textbf{PACS.} PACS is an image dataset for domain generalization \citep{li2017deeperbroaderartierdomain}. It contains 7 content categories, including dog, elephant, giraffe, guitar, horse, and person. Within each category, there are 4 different picture domains, including photo, art painting, cartoon and sketch.

\textbf{CelebA.} CelebA is a large-scale dataset of celebrity face images with 40 labeled attributes \cite{liu2015faceattributes}. In our experiments, we treat the combination of ``gender'' and ``smile'' as the \textit{content} label. The rest of the attributes will be considered as \textit{style}. In the OOD classification experiments, we use hair color as the anchor attribute for the assignment of training and testing splits.

\textbf{Camelyon17-WILDS.} Camelyon17-WILDS is a patch-based variant of the original dataset \citep{bandi2018detection, koh2021wilds}, consisting of histopathology image patches from five hospitals. Notable style variations are evident in the staining, while additional variability can arise from differences in patient populations and image acquisition protocols. The task is binary classification to detect tumor tissue, with an emphasis on generalizing to data from hospitals not seen during training.

\begin{table*}[h]
    \centering
    \caption{\textit{Content} and \textit{style} in each experimental dataset. The \textbf{bold} features are utilized as the anchor attribute in the quantitative evaluation.}
    \label{tab:dataset-def}
    \resizebox{0.9\columnwidth}{!}{
    \begin{tabular}{llcc}
    \toprule
    & modality & content label $(y)$ & style \\
    \midrule
    Styled/Colored-MNIST & $(1\times 28 \times 28)$ & digit & \textbf{style/color} \&  handwriting strokes \\
    PACS (resized) & $(3\times 64 \times 64)$ & object category & \textbf{art domain} \\
    CelebA (resized) & $(3\times 64 \times 64)$ & gender $\times$ smiling & \textbf{hair color} \& remaining characteristics \\
    Camelyon17-WILDS (resized) & $(3\times 64 \times 64)$ & gender $\times$ tumor/normal & \textbf{staining} \& remaining hospital characteristics \\
    \bottomrule
    \end{tabular}
    }
\end{table*}

\begin{figure}[h!]
    \centering
    \begin{minipage}{0.2\linewidth}
        \centering
        \includegraphics[height=3.0cm]{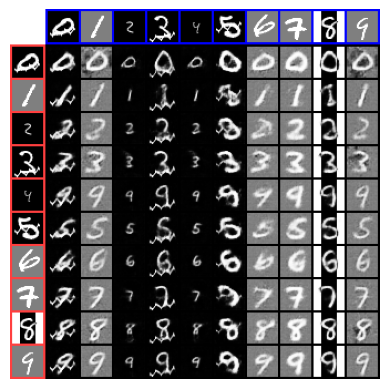}
        \includegraphics[height=3.0cm]{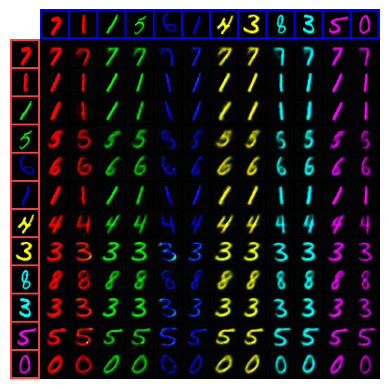}
        \includegraphics[height=3.0cm]{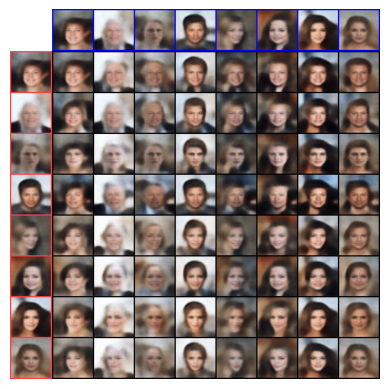}
        \captionsetup{font=scriptsize}
        \caption*{(a) swapping}
    \end{minipage}
    \hspace{0.02\linewidth}
    \begin{minipage}{0.2\linewidth}
        \centering
        \includegraphics[height=3.0cm]{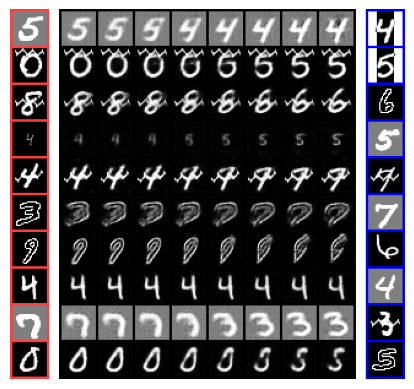}
        \includegraphics[height=3.0cm]{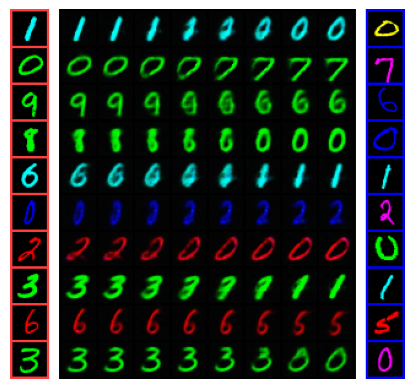}
        \includegraphics[height=3.0cm]{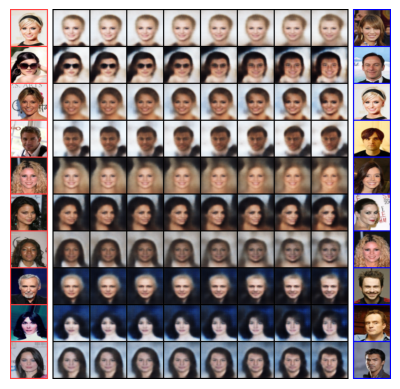}
        \captionsetup{font=scriptsize}
        \caption*{(b) interpolate $\vz^{(c)}$}
    \end{minipage}
    \hspace{0.02\linewidth}
    \begin{minipage}{0.2\linewidth}
        \centering
        \includegraphics[height=3.0cm]{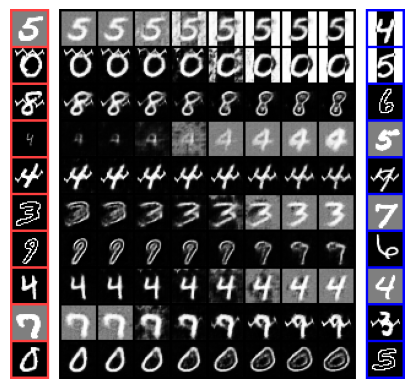}
        \includegraphics[height=3.0cm]{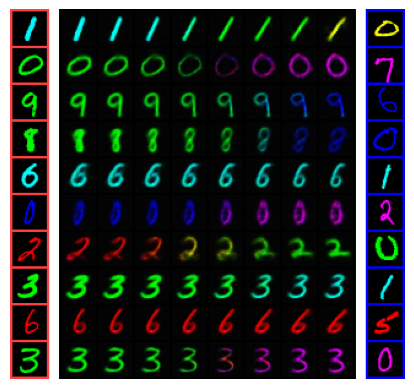}
        \includegraphics[height=3.0cm]{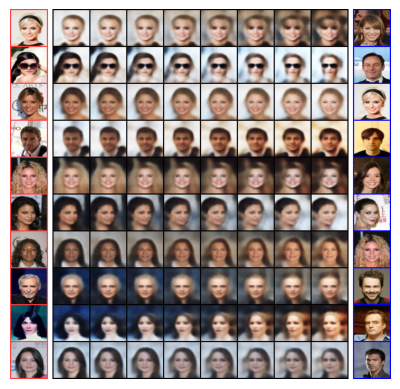}
        \captionsetup{font=scriptsize}
        \caption*{(c) interpolate $\vz^{(s)}$}
    \end{minipage}
    \caption{Swapping and interpolation experiments. In (a), each image is generated using the \textit{content} semantics from the red axis and \textit{style} semantics from the blue axis. (b) interpolates content and (c) interpolates style from the source to the target sample, keeping the other factor fixed.}
    \label{fig:swap-and-inter-mnist}
\end{figure}

\subsection{Swapping and Interpolation}

Fig. \ref{fig:swap-and-inter-mnist} visualizes the swapping and interpolation results from random samples in Styled/Colored-MNIST and CelebA datasets. Appx. \ref{appx:swap-interpolate} provides detailed experiment protocols, and Appx. \ref{appx:more-figs} provides larger graphics and more examples for the comparisons across different methods.

In the swapping experiment results, the diagonal images are the reconstructed samples, and the off-diagonal images are generated from the swapped combinations. The identities for \textit{content} is maintained within each row, and the identities for \textit{style} is consistent within each column. In the interpolation experiment results, either \textit{content} factors or \text{style} factors from the source (left) samples are adjusted to match those of the target (right) samples, while keeping the other attribute unchanged.

From the above analyses, we see that CLEAR enables the model to learn the semantically disentangled latent representations from the data. \textbf{Recall that CLEAR is a weakly supervised framework that does not require style labels, yet it still successfully extracts style characteristics unknown to the model.} During the testing phase, it can perform sample-to-sample conversion through swapping and form local clustering structure without any extra conditional information.

\subsection{Ablation Study}
Appx. \ref{appx:ablation} presents ablation study results that highlight the importance of both contrastive and anti-contrastive regularization terms in the training objective. Their contribution is evidenced by improved clustering patterns in t-SNE visualizations \citep{van2008tsne} and increased gMIG scores.

\subsection{Downstream Classification on OOD samples with shifted style-content associations}

In the classification experiments, the testing sets are all made of samples with OOD \textit{styles}. We design a procedure to evaluate different VAEs' generalizability. Let $\Omega_c = \{c_1,...,c_p\}$ and $\Omega_s = \{s_1,...,s_m\}$ be the set of contents and the set of styles, respectively. For each $c \in \Omega_c$, we only observe $k$ random styles in the training set and evaluate a model's discriminative performance using the other $m-k$ styles in the testing set. Tab. \ref{tab:cls-schema} shows a possible setup used for all datasets except Camelyon17-WILDS, a binary classification dataset for which the same evaluation protocol is inapplicable. Instead, we use its original train-test split, where the test data also OOD.

Averaging performance scores can obscure variability across the random training sets, as different splits may capture different style distributions, leading to varying levels of test difficulty. Hence, we report VAEs' relative improvement from the corresponding CNN baseline classifier for each random split. Appx.~\ref{appx:ood} provides details about the training procedure. We use relative top-1 accuracy and the relative macro averages of one-vs-rest AUROC and AP scores as the metrics for evaluation. For Camelyon17-WILD, we report the standard metric scores.

For clarity, we name $\cL_{\text{SNN}}^{(c)}$-based VAE variants using alternative MI minimization methods by appending the method to CLEAR (e.g., CLEAR-TC, CLEAR-L1OutUB, CLEAR-CLUB-S). The original version using PS is denoted CLEAR-PS. Appx. \ref{appx:mim} provides implementation details of the CLEAR-VAE variants. Table~\ref{tab:camelyon17} presents model performance on OOD test data for the Camelyon17-WILDS task. GVAE and MLVAE show highly variable results, likely due to poor disentanglement as indicated by low gMIG scores during pretraining. Consequently, their downstream classification performance is largely driven by chance.

\begin{table*}[h]
    \centering
    \caption{Model performance on OOD test data from Camelyon17-WILDS.}
    \label{tab:camelyon17}
    \resizebox{\columnwidth}{!}{
    \begin{tabular}{ccccccccc}
    \toprule
    & vanilla CNN & LAM \citep{pmlr-v244-gao24a}& CLEAR-PS \textbf{(ours)} & CLEAR-CLUB-S & CLEAR-L1OutUB & CLEAR-TC & MLVAE & GVAE \\
    \midrule
    Acc. & $0.638~(\pm 0.022)$ & $0.713~(\pm 0.012)$ & $\boldsymbol{0.747~(\pm 0.019)}$ & 
    $0.682~(\pm 0.046)$ & $0.702~(\pm 0.039)$ & $0.72~(\pm 0.024)$ & $0.641~(\pm 0.116)$ & $0.661~(\pm 0.084)$ \\
    AUC  & $0.713~(\pm 0.030)$ & $0.804~(\pm 0.025)$ & $\boldsymbol{0.832~(\pm 0.019)}$ & 
    $0.757~(\pm 0.046)$ & $0.780~(\pm 0.040)$ & $0.788~(\pm 0.036)$ & $0.729~(\pm 0.085)$ & $0.740~(\pm 0.074)$ \\
    AP   & $0.667~(\pm 0.046)$ & $0.791~(\pm 0.012)$ & $\boldsymbol{0.804~(\pm 0.027)}$ & 
    $0.686~(\pm 0.064)$ & $0.719~(\pm 0.055)$ & $0.748~(\pm 0.069)$ & $0.682~(\pm 0.121)$ & $0.747~(\pm 0.093)$ \\
    \bottomrule
    \end{tabular}
    }
\end{table*}

\begin{figure}[h]
    \centering
    \begin{minipage}{\linewidth}
        \centering
        \includegraphics[width=0.7\linewidth]{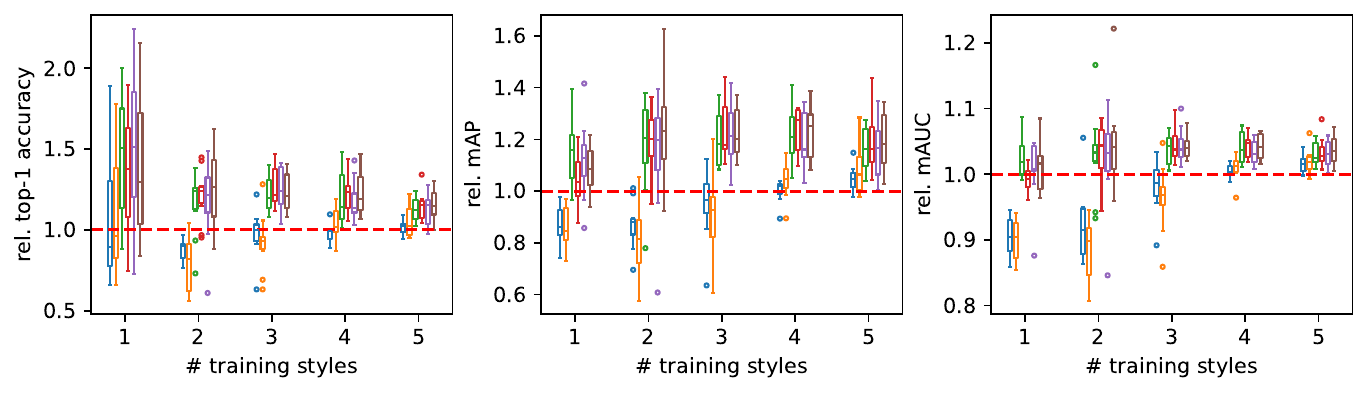}
        \vspace{-0.25cm}
        \captionsetup{font=scriptsize}
        \caption*{(a) Styled-MNIST}
        \label{fig:bplot-mnist}
    \end{minipage}

    \begin{minipage}{\linewidth}
        \centering
        \includegraphics[width=0.7\linewidth]{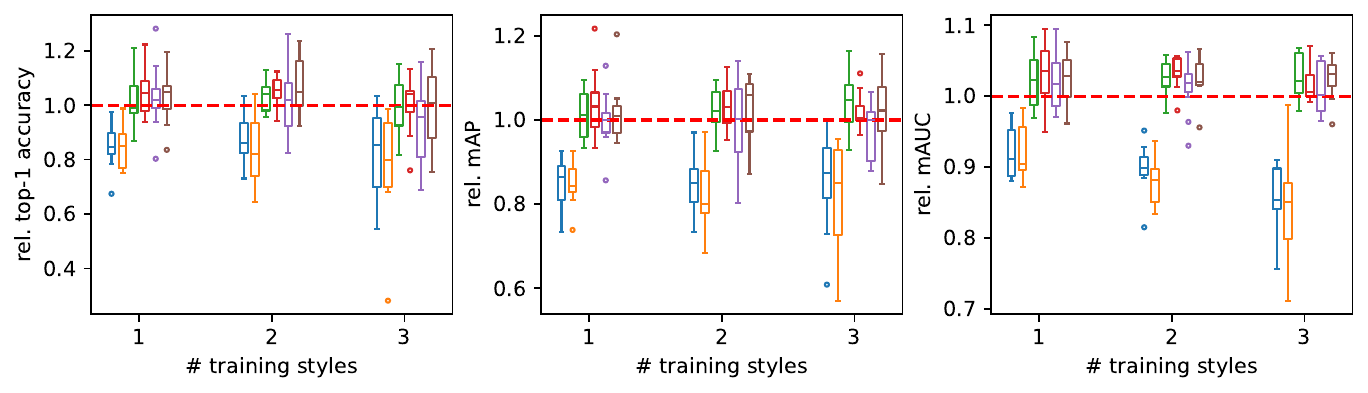}
        \vspace{-0.25cm}
        \captionsetup{font=scriptsize}
        \caption*{(b) PACS}
        \label{fig:bplot-pacs}
    \end{minipage}
    
    \begin{minipage}{\linewidth}
        \centering
        \includegraphics[width=0.7\linewidth]{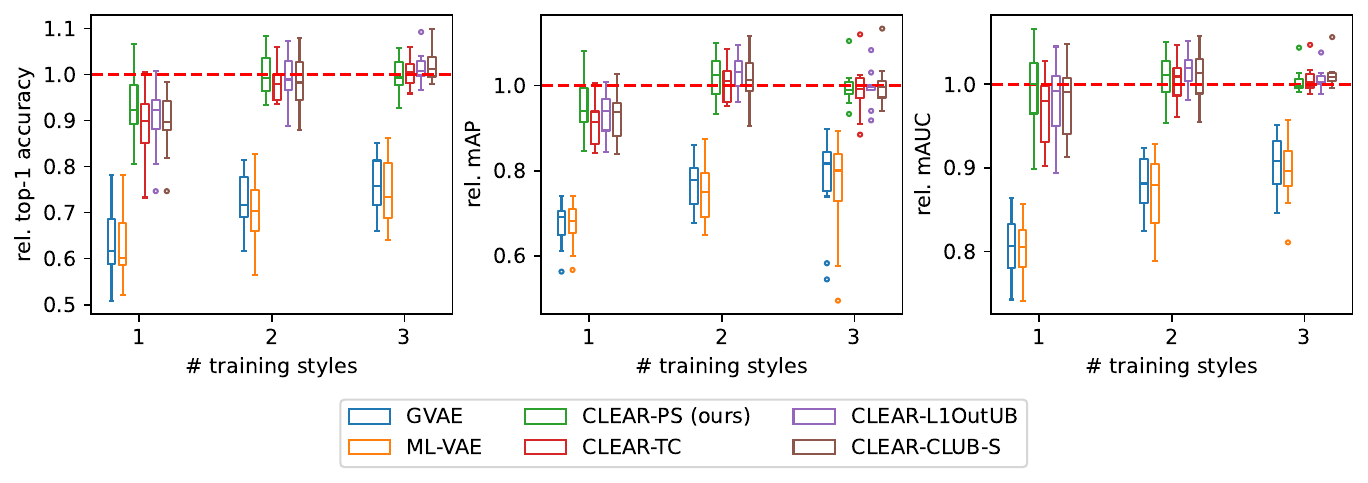}
        \vspace{-0.25cm}
        \captionsetup{font=scriptsize}
        \caption*{(c) CelebA}
        \label{fig:bplot-celeba}
    \end{minipage}

    \caption{Downstream classifications on OOD samples. CNN baselines achieve descent performance when content-style distributions largely overlap between training and testing set.}
    \label{fig:downstream-bplots}
\end{figure}

Fig. \ref{fig:downstream-bplots} visualizes different models' relative performances across 10 different random splits. Contrastive learning based VAE encoders in general outperform ML-VAE and GVAE. All these variants can more effectively disentangle $\vz^{(c)}$ and $\vz^{(s)}$ from each other on the unseen combinations. Among the CLEAR variants, the PS-based approach -- with a simpler training framework -- achieves performance comparable to that of the other MI minimization methods.

When more styles are observed at training time, the baseline models' absolute performance increases. There exists variability in the relative performance, because the training and testing data are randomly generated. As the total number of training styles increases, the overlap between samples' style feature space increases, and the benefit of using disentangled representations stabilizes.

\section{Conclusion \& Limitation \& Broader Impact}

We propose and analyze CLEAR, a weakly supervised framework for learning semantically disentangled representations that relies only on content labels to define contrastive pairs. We then conduct extensive swapping and interpolation experiments using CLEAR-VAE, which provide compelling evidence that CLEAR can effectively disentangle and recognize both \textit{content} and \textit{style} features. Importantly, our classification experiments demonstrate that this disentanglement can enhance downstream prediction performance on content-style combinations not observed during training, which has important implications for algorithmic generalizability and fairness more broadly. 

A limitation of our current work is that the CLEAR framework has not yet been scaled to other probabilistic generative models or extended to multi-modal settings. This is an important direction for future work. Further, while supervised contrastive losses such as SupCon and our modified SNN are based on the InfoNCE objective with multiple positive pairs per anchor, their theoretical connection to MI has not been rigorously established. Our empirical results suggest that minimizing $\mathcal{L}_{\text{SNN}}$ promotes MI maximization, but a closed-form relationship still remains unclear.

The disentanglement achieved by CLEAR not only enhances the interpretability of representation learning, but also yields content features that are resilient to superficial style variations. With respect to societal impact in the context of healthcare, we provide a promising approach to isolate meaningful pathological findings from patient-specific factors, including race and gender, as well as hospital-specific factors, such as data collection and imaging protocols. CLEAR can be used to help models focus on clinically relevant signals for all patients rather than spurious, cohort-specific associations, thereby improving models' fairness and external validity.

\newpage

{
\small

\bibliographystyle{plainnat}
% \bibliography{clear/neurips2025}

}

%%%%%%%%%%%%%%%%%%%%%%%%%%%%%%%%%%%%%%%%%%%%%%%%%%%%%%%%%%%%
\newpage
\appendix

\setcounter{equation}{0}
\renewcommand{\theequation}{\thesection.\arabic{equation}}
\renewcommand{\thefigure}{\thesection.\arabic{figure}}
\renewcommand{\thetable}{\thesection.\arabic{table}}
\section{ELBO}
\subsection{ELBO Decomposition}
\label{appx:elbo-decomp}

Assume $\vz^{(c)}$ and $\vz^{(s)}$ are independent \textit{a priori} and \textit{a posteriori}. Following the Markov graphical model in Fig. \ref{fig:clear-vae}, we can derive an ELBO as follows.

\begin{align}
    \log p_{\vtheta}(\vx) 
    &= \log \iint p_{\vtheta}(\vx, \vz, y) \, \frac{q_{\vphi}(\vz | \vx)}{q_{\vphi}(\vz | \vx)} \, d\vz \, dy \\
    &= \log \int p(y) \left( \int p_{\vtheta}(\vx, \vz | y) \, \frac{q_{\vphi}(\vz | \vx)}{q_{\vphi}(\vz | \vx)} \, d\vz \right) dy \\
    &\overset{\text{(Jensen)}}{\geq} \E_{y \sim p(y)} \left[ \log \int q_{\vphi}(\vz | \vx) \, \frac{p_{\vtheta}(\vx, \vz | y)}{q_{\vphi}(\vz | \vx)} \, d\vz \right] \\
    &\overset{\text{(Jensen)}}{\geq} \E_{y \sim p(y)} \E_{\vz \sim q_{\vphi}(\vz | \vx)} \left[ \log \frac{p_{\vtheta}(\vx, \vz | y)}{q_{\vphi}(\vz | \vx)} \right] \\
    &= \E_{y \sim p(y)} \E_{\vz \sim q_{\vphi}(\vz | \vx)} \left[ \log p_{\vtheta}(\vx | \vz) + \log p(\vz | y) - \log q_{\vphi}(\vz | \vx) \right] \\
    &= \E_{y \sim p(y)} \E_{\vz \sim q_{\vphi}(\vz | \vx)} \Big[ 
        \log p_{\vtheta}(\vx | \vz) 
        + \log p(\vz^{(c)} | y) + \log p(\vz^{(s)}) \notag \\
    &\qquad\qquad\qquad\qquad - \log q_{\vphi}(\vz^{(c)} | \vx) - \log q_{\vphi}(\vz^{(s)} | \vx) 
    \Big]
\end{align}

Moreover,
\begin{equation}
    p(\vz^{(c)}|y) = \frac{f(y|\vz^{(c)})p(\vz)}{p(y)}
\end{equation}

Therefore,
\begin{align}
    \log p_{\vtheta}(x) &\ge \E_{\vz\sim q_{\vphi}}[\log p_{\vtheta}(\vx|\vz)] 
    - D_{\text{KL}}\Big(q_{\vphi}(\vz^{(c)}|\vx) \| p(\vz^{(c)})\Big)
    - D_{\text{KL}}\Big(q_{\vphi}(\vz^{(s)}|\vx) \| p(\vz^{(s)})\Big) \notag  \\
    &\quad + \E_{y\sim p,z^{(c)}\sim q_{\vphi}}\left[\log \frac{f(y|\vz^{(c)})}{p(y)}\right]
\end{align}

\subsection{Predictive Interpretation}
\setcounter{figure}{0}
\begin{align}
    \log p_{\vtheta}(x)
    &\ge \E_{\vz\sim p_{\vphi}}[\log p_{\vtheta}(\vx|\vz)] - D_{\text{KL}}\Big(q_{\vphi}(\vz^{(c)}|\vx) \| p(\vz^{(c)})\Big) - D_{\text{KL}}\Big(q_{\vphi}(\vz^{(s)}|\vx) \| p(\vz^{(s)})\Big) \notag \\
    &\quad + \E_{y\sim p}\E_{z\sim q_{\vphi}}[\log f(y|\vz^{(c)})] - H(y)
    \label{eq:elbo-pred}
\end{align}

Here maximizing $\log_{\vtheta}(\vx)$ encourages maximizing the predictive-ness of $y$ using $\vz^{(c)}$. To improve the predictive-ness, we can either have an adversarial classifier or use supervised contrastive learning. The minimization objective must include the following parts: reconstruction error, KL divergences for $\vz^{(c)}$ and $\vz^{(s)}$, and a supervised contrastive regularization.

\subsection{Mutual Information Interpretation}

\subsubsection{Connection with InfoNCE}
\label{appx:elbo-mi}
Following Eqn. (\ref{eq:elbo-pred}), we can combine the last two terms to form. Unfortunately, it is not the mutual information, since MI is integrated w.r.t. the joint distribution.

\begin{equation}
    \E_{y\sim p,\vz^{(c)}\sim q_{\vphi}} \Bigg[ \log \frac{f(y|\vz^{(c)})}{p(y)} \Bigg] \ne I(\vz^{(c)};y)
\end{equation}

However, in the Monte Carlo-1 (MC-1) estimation where we only sample $\vz^{(c)}$ once, the notion of maximizing the log-ratio makes it equivalent to maximizing InfoNCE loss. This in turn encourages maximizing $I(\vz^{(c)};y)$.

\begin{equation}
    \max \E_{y\sim p,\vz^{(c)}\sim q_{\vphi}} \Bigg[ \log \frac{f(y|\vz^{(c)})}{p(y)} \Bigg]
    \Leftrightarrow \min \frac{1}{N}\sum_{i=1}^{N} -\log \frac{h(y_i,\vz_i^{(c)})}{\sum_{y_j\in\mathcal{Y}}h(y_j, \vz_i^{(c)})} \Leftrightarrow \max I(y; \vz^{(c)})
\end{equation}

Therefore, the MLE becomes
\begin{align}
    \max_{\vtheta, \vphi} \; \log p_{\vtheta}(\vx) 
    &\;\Leftrightarrow\; \max_{\vtheta, \vphi} \Big(  
        \E_{\vz \sim q_{\vphi}(\vz | \vx)} [\log p_{\vtheta}(\vx | \vz)] \notag \\
    &\quad - D_{\mathrm{KL}} \big( q_{\vphi}(\vz^{(c)} | \vx) \,\|\, p(\vz^{(c)}) \big) 
          - D_{\mathrm{KL}} \big( q_{\vphi}(\vz^{(s)} | \vx) \,\|\, p(\vz^{(s)}) \big) \notag \\
    &\quad + I(y ; \vz^{(c)}) \Big)
\end{align}

\subsubsection{From InfoNCE to SNN}
\label{appx:snn-derivation}
Now, let's take a an arbitrary positive pair $(y_i=k, \vz^{(c)}_i)$. Assume the log-bilinear model $h(y_i, \vz^{(c)}_i)$ is parameterized a learnable and flexible weight $W$. Then, we extend the log-bilinear model:
\begin{equation}
    h(y_i, \vz_i^{(c)}) = \exp(\vec{y}_k^{\top}W\vz^{(c)}_i) = \exp\left(\langle\hat{\ve}_k,\hat{\vz}^{(c)}\rangle/\tau\right) =
    \exp\left(\mathrm{sim}(\hat{\ve}_k, \hat{\vz}^{(c)})/\tau\right)
\end{equation}
where $\vec{y}$ is the one-hot encoding, $\hat{\boldsymbol{e}}_k$ is normalized class embedding for class $k$, $\hat{\vz}^{(c)}_i$ is the normalized $\vz^{(c)}_i$, and $\tau$ is the temperature. Since $\hat{\boldsymbol{e}_k}$ is the class embedding, let $\hat{\boldsymbol{e}}_k$ be the average of the normalized latent representation for samples with class label $y=k$. Specifically,
\begin{equation}
    \hat\ve_k = \frac{1}{n_k}\sum_{y_j=k}\hat{\vz}_j^{(c)}
\end{equation}
and $n_k$ is the class size in the data batch. We use inner product for the simplicity of notation, which is the same as cosine similarity for normalized vectors.

Since multiple observations can share the same value for the content label $y$, we would like to contrast $\vz^{(c)}$ with all of these instances. Then, we obtain a predecessor of $\mathcal{L}_{\text{SNN}}^{(c)}$, where the log-bilinear model is used in place of the cosine similarity.
\begin{equation}
\tilde\cL^{(c)}_{\text{SNN}} = \frac{1}{N} \sum_{i=1}^{N} -\log \frac{\sum_{y_j=y_i}h(y_j, \vz_i^{(c)})}{\sum_{j}h(y_j, \vz_i^{(c)})}
\end{equation}
The MC-1 estimator of $\cL_{\text{InfoNCE}}^{(c)}$ in Eqn. \ref{eqn:infonce} over the observed data distribution is:
\begin{equation}
    \cL_{\text{InfoNCE}}^{(c)} = \frac{1}{N} \sum_{i=1}^{N} -\log \frac{h(y_i, \vz_i^{(c)})}{\sum_{j}h(y_j, \vz_i^{(c)})}
\end{equation}
Then, we have $\cL_{\text{InfoNCE}}^{(c)} \ge \tilde\cL^{(c)}_{\text{SNN}}$ and
in \ref{appx:big-oh-approx}, we show that $\cL_{\text{SNN}}^{(c)} > \tilde\cL^{(c)}_{\text{SNN}}$. Although $\cL_{\text{InfoNCE}}^{(c)} \ge -I(y,\vz^{(c)})$, the analytical relationship between $\cL_{\text{SNN}}^{(c)}$ and $I(y,\vz^{(c)})$ still remain puzzling. However, numerical experiment for Fig. \ref{fig:mi-simulation} (a) empirically verify that minimizing $\cL_{\text{SNN}}^{(c)}$ encourages to increase the MI.

\subsection{PS losses are MI upper bound}
\label{ps-mi-ub}
Recalling the pair-switched InfoNCE loss from Eqn. \ref{eqn:ps-infonce}, its MC-1 estimate is:
\begin{align}
    &\quad~~\frac{1}{N} \sum_{i=1}^{N}-\log \frac{\sum_{j \ne i}h(y_j, \vz^{(s)}_i)}{\sum_{j}h(y_j, \vz^{(s)}_i)} \\
    &=\frac{1}{N} \sum_{i=1}^{N}\log \left(1 + \frac{h(y_i, \vz_i^{(s)})}{\sum_{j}h(y_j, \vz^{(s)}_i)} \right) \\
    &=\frac{1}{N} \sum_{i=1}^{N}\log \Bigg( 1 + \frac{h(y_j, \vz^{(s)}_i)}{(N-1) \underbrace{\E_{y}[h(y, \vz_i^{(s)})]}_{=1}} \Bigg) + o_p(1) \\
    &\ge \frac{1}{N} \sum_{i=1}^{N} \log(h(y_j, \vz^{(s)}_i)) - \log(N-1) + o_p(1) \\
    &\ge I(y, \vz^{(s)}) - \log(N)
\end{align}

Similarly, we remove all $y_j$ with value same as $y_i$ (which include more than just one pair) from the numerator. Then, we obtain $\tilde\cL_{\text{PS-SNN}}^{(s)}$
\begin{equation}
    \tilde\cL_{\text{PS-SNN}}^{(s)}
    = \frac{1}{N} \sum_{i=1}^{N} - \log \frac{\sum_{y_j\ne y_i}h(y_j, \vz^{(s)}_i)}{\sum_{j}h(y_j, \vz^{(s)}_i)} \ge \cL_{\text{PS-InfoNCE}}^{(s)} \ge I(y, \vz^{(s)}) - \log(N)
\end{equation}

In \ref{appx:big-oh-approx}, we show that $\cL_{\text{PS-SNN}}^{(s)} > \tilde\cL^{(s)}_{\text{PS-SNN}}$. Therefore,

\begin{equation}
    \cL_{\text{PS-SNN}}^{(s)} > I(y, \vz^{(s)}) - \log(N)
\end{equation}

\begin{figure}[h]
    \centering
    \resizebox{0.4\linewidth}{!}{\begin{tikzpicture}
    \fill[blue!20] (-1,-0.2) rectangle (5.1,0.2);
    % Draw the horizontal axis
    \draw[->] (-3,0) -- (5,0) node[right] {};

    % Draw ticks and labels below the axis
    % Draw ticks and labels below the axis (removed x = –3)
    \foreach \x/\label in {
        2/{$\mathcal{L}_{\text{PS-InfoNCE}}^{(s)}$},
        -1/{\textcolor{blue}{$-\log(N)$}},
        0/{\textcolor{blue}{0}}
    } {
        \draw (\x,0.1) -- (\x,-0.1);
        \node[below, yshift=-0.15cm] at (\x,0) {\label};
    }
    
    % Draw ticks and labels above the axis (removed x = –2)
    \foreach \x/\label in {
         0.85/{\textcolor{red}{$I - \log(N)$}},
         2.5/{$\mathcal{L}_{\text{PS-SNN}}^{(s)}$}
    } {
        \draw (\x,0.1) -- (\x,-0.1);
        \node[above, yshift=0.15cm] at (\x,0) {\label};
    }
    
    % mark I-log(N) and 0
    \filldraw[red] (0.85,0) circle (2pt);
    \filldraw[blue] (0,0) circle (2pt);

    % mark I-log(N) and 0
    \filldraw[red] (0.85,0) circle (2pt);
    \filldraw[blue] (0,0) circle (2pt);
\end{tikzpicture}}
    \caption{Relationship between MI and different losses}
    \label{fig:ps-ineq}
\end{figure}
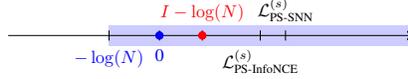

Fig. \ref{fig:ps-ineq} illustrates the $I-\log(N)$ and various loss functions on a number line, with the shaded interval $[\log(N), +\infty)$ as the range of $I-\log(N)$. Its ideal optimal is $-\log(N)$ when $I = 0$. Although $-\cL_{\text{SNN}}^{(s)}$ may fall within the interval of $(-\log(N), 0)$ and and may exhibit the closest proximity to the optimal value, this relationship is not guaranteed to hold consistently during the optimization process. However, based on the aforementioned discussion, $\cL_{\text{PS-SNN}}^{(s)}$ offers better numerical stability.

\subsection{Simplifying $h(\cdot)$ to $\text{sim}(\cdot)$}
\label{appx:big-oh-approx}

In the following derivation we generic $\vz$ to denote both $\vz^{(c)}$ and $\vz^{(s)}$. Following the previous notation and WLOG assuming $y_j = k$ and $\{k_l\}_{l=1}^{n_k}$ is the index sequence for samples with content labeling being $k$.
\begin{align}
    h(y_j, \vz_i) &= \exp(\vec{y}_kW\vz_i) \\
    &= \exp\left(\frac{1}{n_k} \sum_{l} \langle \vz_{k_l}, \vz_i \rangle\right) \\
    &\le \frac{1}{n_k} \sum_l \exp\left(\frac{1}{\tau} \langle \vz_{k_l}, \vz_i \rangle\right)
\end{align}

Now, assume $\frac{1}{\tau} \langle \vz_{k_l}, \vz_i \rangle = m_{ki} + \delta_{k_li}$, where $m_{ki}$ is the average scaled inner product between representations of $k_l$-th and $i$-th samples and $\delta_{k_li}$ is the fluctuation that sum up to 0 over $\{k_l\}$. Then, we can derive a Taylor expansion on $h(y_i, \vz_i)$ as the follows.

\begin{align}
    \frac{1}{n_k} \sum_{l=1}^{n_k} \exp(m_{ki} + \delta_{k_li})
    &= \exp(m_{ki}) \left[\frac{1}{n_k} \sum_{l=1}^{n_k} (1 + \delta_{k_li} + \frac{1}{2}\delta_{k_li}^2 + O(\delta_{k_li}^3))\right] \\
    &= \exp(m_{ki}) \left[1 + \frac{1}{2}\Delta_{ki}^2 + O(\delta_{k_li}^3) \right]
\end{align}
where $\Delta_{ki}^2 = \text{Var}(\delta_{k_li})$. Thus,

\begin{equation}
    \frac{1}{n_k} \sum_l \exp\left(\frac{1}{\tau}\langle \vz_{kl}, \vz_i\rangle\right) = h(y_i, \vz_i)(1 + O(\Delta_{ki}^2))
\end{equation}

Also note that $\log(A + O(B)) = \log(A) + O\left(\frac{A}{B}\right)$. Therefore, the $\cL_{\text{SNN}}^{(c)}$ on the $i$-th observation (\ref{eqn:approx-snn}) is:

\begin{align}
    -\log \frac{\text{pos}_i}{\text{pos}_i + \text{neg}_i} 
    &= -\log \frac{\sum_{l}h(y_{k_l},\vz_i^{(c)}) + O(\Delta_{ki}^2)}{\sum_{k',l}h(y_{k'_l},\vz_i^{(c)}) + \sum_{k'}O(\Delta_{k'i}^2)} \\
    &\ge -\log \frac{\sum_{l}h(y_{k_l},\vz_i^{(c)})}{\sum_{k',l}h(y_{k'_l},\vz_i^{(c)})} 
\end{align}

Similarly, for PS-SNN

\begin{align}
    -\log \frac{\text{neg}_i}{\text{pos}_i + \text{neg}_i} 
    &= -\log \frac{\sum_{k',l}h(y_{k'_l},\vz_i^{(s)}) + \sum_{k'\ne k}O(\Delta_{k'i}^2)}{\sum_{k',l}h(y_{k'_l},\vz_i^{(s)}) + \sum_{k'}O(\Delta_{k'i}^2)} \\
    &\ge -\log \frac{\sum_{k'l}h(y_{k'_l},\vz_i^{(s)})}{\sum_{k',l}h(y_{k'_l},\vz_i^{(s)})}
\end{align}

\section{Simulation Study Setup}
\label{appx:mi-simulation}
Assume $\vz$ is sampled from a Mixture of Gaussians. Let $y \sim \text{Multinomial}(3,\frac{1}{3},\frac{1}{3}, \frac{1}{3})$, and $\vz|y \sim \mathcal{N}(\mu_j, \Sigma_j)$ and $j \in \{1,2,3\}$. Due to the lack of closed-form solution to calculate $I(y;\vz)$ analytically, we use the KNN estimator \citep{ross2014mutual} (with $k=3$) to numerically estimate it. According to simulation results from \citet{ross2014mutual}, the estimate has small bias and variability when the sample size is large enough.

To simplify the simulation setup, we use isotropic Gaussian distributions with the same variance such that $\Sigma_j = \sigma^2 \mathbf{I}$. Then, we can easily grow $I(y;\vz)$ by decreasing $\sigma^2$ or reduce $I(y;\vz)$ by increasing $\sigma^2$. To recreate Fig. \ref{fig:mi-simulation}, we sample $\{(y_i, \vz_i)\}_{i=1}^{N}$ at each step and calculate the KNN estimate and $\cL_{\text{SNN}}$/$\cL_{\text{PS-SNN}}$. After each 100 steps, we decrease/increase $\sigma^2$. We use following parameter configuration to generate data at each step: $N = 1500, \mu_1 = (-1,-1,-1), \mu_2=(2,2,2), \mu_2=(2,2,2)$, and $\sigma \in \{1,1.3,...,4\}$ (Of note, we are manipulating the SD in practice).

\newpage

\section{Swapping \& Interpolation Experiment Procedure}
\label{appx:swap-interpolate}
\textbf{Content and style swapping.} We assess the quality of representation learning by manipulating $\vz^{(c)}$ and $\vz^{(s)}$ from testing samples. We follow the ``swapping'' experiment protocol proposed by \citet{mathieu2016disentangling} and \citet{bouchacourt2018mlvae}. Specifically, we extract the latent representations $\vz^{(c)}$ and $\vz^{(s)}$ for both $\vx_i$ and $\vx_j$, swap either the content representation or style representation between the samples, and finally generate a new sample by decoding the representation that combines the content representation from one sample with the style representation from another. In the experimental results, we present a grid of generated samples. Each row corresponds to a fixed content representation $\vz^{(c)}$, and each column corresponds to a fixed style representation $\vz^{(s)}$. The image at coordinate $(i,j)$ in the grid is generated using the content latent vector $\vz_i^{(c)}$ and the style latent vector $\vz_j^{(s)}$.

\textbf{Interpolation.} We also perform interpolation analysis on the latent representations to evaluate the quality of generated data. We generate sequences of images along the line segments between representations. To investigate disentanglement between \textit{content} factors and \textit{style} factors, we fix one representation and interpolate the other. For instance, when interpolating along the \textit{style}, we generate images along the line segment between $(\vz^{(c)}_i, \vz^{(s)}_i)$ and $(\vz^{(c)}_i, \vz^{(s)}_j)$. When interpolating along \textit{content}, we generate images along the line segment between $(\vz^{(c)}_i, \vz^{(s)}_i)$ and $(\vz^{(c)}_j, \vz^{(s)}_i)$.

\section{OOD Classification Tasks Setup}
\label{appx:ood}
\begin{table*}[h]
    \caption{One possible realization of the train-test split setup for OOD sample classification, where the training and testing styles are randomly assigned.}
    \centering
    \label{tab:cls-schema}
    \resizebox{0.7\linewidth}{!}{
    \begin{tabular}{cccccccc}
        \toprule
        & \multicolumn{2}{c}{$k=1$} & \multicolumn{2}{c}{$k=2$} & $\cdots$ & \multicolumn{2}{c}{$k=m-1$} \\
        \cmidrule(lr){2-3} \cmidrule(lr){4-5} \cmidrule(lr){7-8}
         & train & test & train & test & $\cdots$ & train & test \\
        \midrule
        $c_1$ & $\{s_1\}$ & $\Omega_s \setminus \{s_1\}$ & $\{s_3, s_5\}$ & $\Omega_s \setminus \{s_3, s_5\}$     & $\cdots$ & $\Omega_s \setminus \{s_2\}$  & $\{s_2\}$ \\
        $c_2$ & $\{s_4\}$ & $\Omega_s \setminus \{s_4\}$ & $\{s_1, s_2\}$ & $\Omega_s \setminus \{s_1, s_2\}$     & $\cdots$ & $\Omega_s \setminus \{s_1\}$  & $\{s_1\}$ \\
        $\vdots$  & & & & & $\cdots$ & & \\
        $c_p$ & $\{s_1\}$ & $\Omega_s \setminus \{s_1\}$ & $\{s_3, s_4\}$ & $\Omega_s \setminus \{s_3, s_4\}$     & $\cdots$ & $\Omega_s \setminus \{s_m\}$  & $\{s_m\}$ \\
        \bottomrule
    \end{tabular}
    }
\end{table*}

We use a two-step process to obtain downstream predictions from a VAE. First, we pre-train the VAE model and freeze its weights. Next, we extract the latent representation, $\vmu^{(c)}$, and feed it into a two-layer MLP. To ensure fair comparisons between VAEs and baseline CNN models, all VAE encoders share the same architecture as the CNN baseline.

\section{Mutual Information gap between groups of latent variables}
\label{sec:gmig}

\citet{chen2018isolating} proposed a interpretable, classifier-free metric based on $I(z_{j};y) / H(y)$, the normalized MI between a latent variable $z_{j}$ and a ground truth factor $y$. For the metric calculation, we can use a more efficient nonparameteric nearest-neighbor approach to directly estimate the MI \citep{ross2014mutual}. The complete metric for a label $y$, known as the Mutual Information Gap (MIG), is defined as the difference between the top two latent variables with the highest normalized MI \cite{chen2018isolating}. Whereas the original definition is an average over all possible ground truth labels, in our scenario, we consider MIG only for the class label $y$:
\begin{equation}
    \text{MIG}(y) = \frac{1}{H(y)} \left( I(z^*;y) - \max_{z_j\ne z^*} I(z_{j};y) \right)    
\end{equation}
where $z^{*} = \text{argmax}_z~I(z;y)$. The MIG$(y)$ quantifies the degree of disentanglement at level of individual latent variables. We adapt it to our specific scenario, where $\vz = (\vz^{(c)}, \vz^{(s)})$, as follows:
\begin{equation}
\text{gMIG}(y) = \frac{1}{H(y)} \left( \frac{1}{d_c} \sum_{j=1}^{d_c} I(z_j^{(c)};y) - \frac{1}{d_s} \sum_{j=1}^{d_s} I(z_j^{(s)};y) \right)
\end{equation}

We call this modified version the group Mutual Information Gap (gMIG). Bounded between -1 and 1, it quantifies the gap between $\vz^{(c)}$'s average association to $y$ and $\vz^{(s)}$'s average association to $y$.

\section{Hyperparameters' Effects on the Disentanglement in CLEAR}
\label{hyperparam-effect}

The $\beta$-VAE promotes disentanglement across individual latent variables when $\beta > 1$ \citep{higgins2017beta}. Without supervision, the model is incapable of allocating content relevant information in the designated partition $\vz^{(c)}$. The individual latent variable's association to the content label will be canceled out when we calculate the group average in gMIG. Therefore, we would expect to see almost 0 gMIG between $\vz^{(s)}$ and $\vz^{(c)}$ in $\beta$-VAE.

\begin{figure}[h!]
    \centering
    % First plot (top)
    \begin{minipage}{\linewidth}
        \centering
        \includegraphics[width=\linewidth]{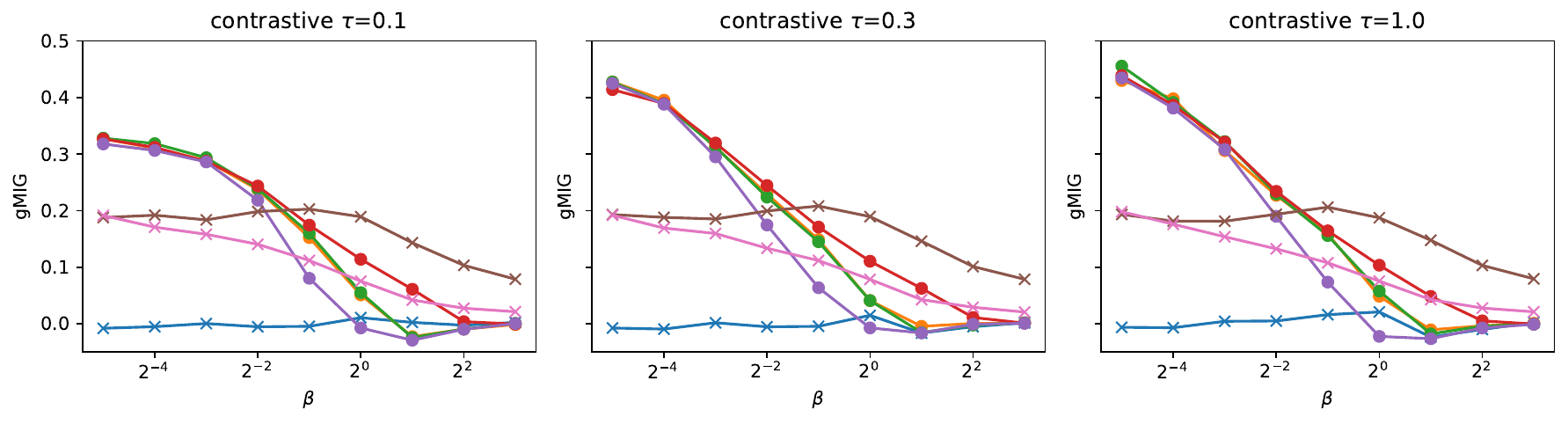}
        \vspace{-0.7cm}
        \captionsetup{font=scriptsize}
        \caption*{(a) Styled-MNIST}
        \label{fig:gmig-mnist}
    \end{minipage}
    
    % Second plot (bottom)
    \begin{minipage}{\linewidth}
        \centering
        \includegraphics[width=\linewidth]{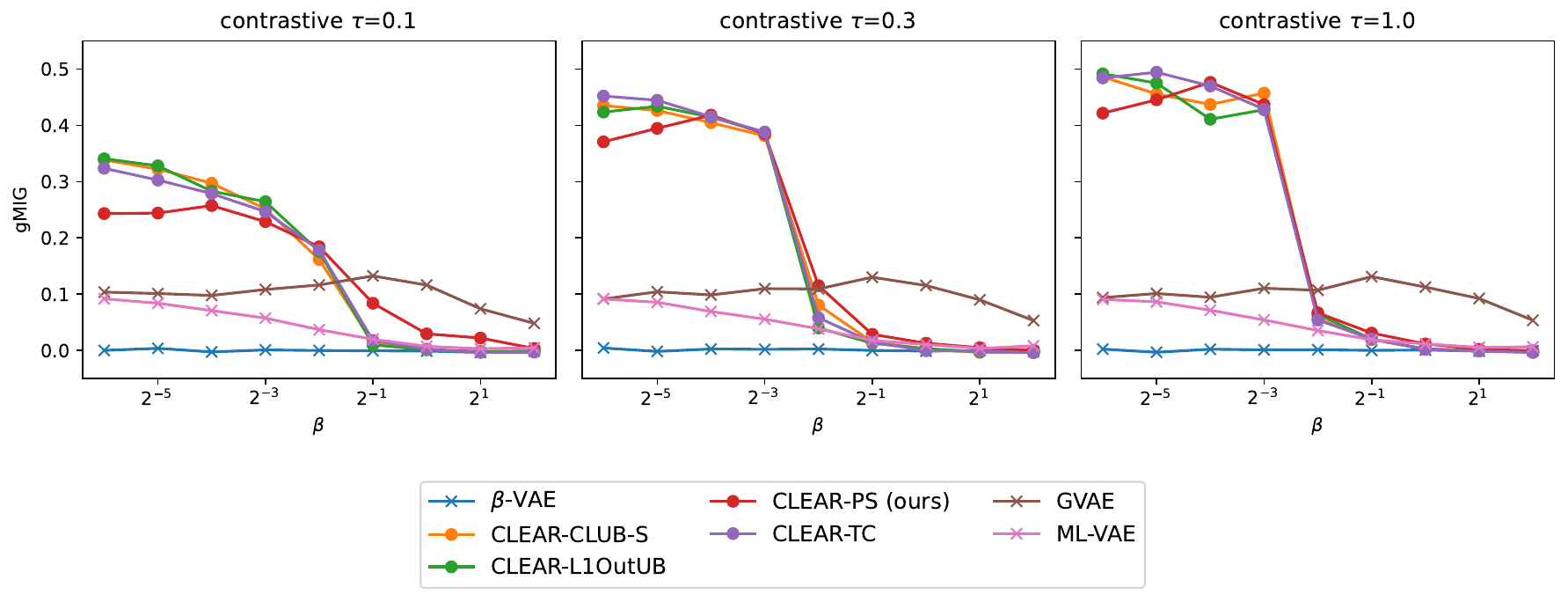}
        \vspace{-0.7cm}
        \captionsetup{font=scriptsize}
        \caption*{(b) CelebA}
        \label{fig:gmig-celeba}
    \end{minipage}

    \vspace{-0.25cm}
    \caption{gMIG changes with respect to $\beta$ and $\tau$, averaged over three random seeds}
    \label{fig:gmig-plots}
\end{figure}

On the contrary, all other VAE variants achieve significant levels of disentanglement between $\vz^{(s)}$ and $\vz^{(c)}$ when $\beta < 1$. For CLEAR-based variants, gMIG is limited when $\beta > 1$. This suggests that we have to avoid complete factorization in latent variables to achieve disentanglement between \textit{style} and \textit{content}. Fig.~\ref{fig:gmig-plots} visualizes this effect. To achieve a disentanglement at the semantic level, we have to allow an appropriate level of entanglement at the individual latent factor level. 

VAE variants with contrastive regularization tend to achieve higher gMIG with smaller values of $\beta$ and larger values of $\tau$. However, the numerical gain in gMIG comes at the expense of poorer representation learning by other measures. With a small $\beta$, the VAE focuses on reconstructing the training data, thereby having reduced ELBO and limited ability to generate diverse samples \citep{alemi2018fixing}. With a large $\tau$, the model can be insensitive to important details (e.g. zigzag background vs. identity background). Moreover, a large $\tau$ (e.g. $\tau = 1$) occasionally causes numerical instability error in CLEAR-TC's discriminator. We would have to re-initialize the model and restart the training.

\begin{table*}[h]
    \centering
    \caption{gMIG comparison across different metrics used in contrastive loss for CLEAR-VAE}
    \label{tab:sims-metric}
    \resizebox{0.75\linewidth}{!}{
    \begin{tabular}{cclclclcl}
        \toprule
        & \multicolumn{2}{c}{Styled-MNIST} & \multicolumn{2}{c}{Colored-MNIST} & \multicolumn{2}{c}{CelebA} & \multicolumn{2}{c}{PACS} \\
        \cmidrule(lr){2-3} \cmidrule(lr){4-5} \cmidrule(lr){6-7} \cmidrule(lr){8-9}
        & gMIG & $(d_{\vz}, \alpha, \beta, \tau)$ & gMIG & $(d_{\vz}, \alpha, \beta, \tau)$ & gMIG & $(d_{\vz}, \alpha, \beta, \tau)$ & gMIG & $(d_{\vz}, \alpha, \beta, \tau)$ \\
        \midrule
        \textbf{Cosine Similarity} & \textbf{0.31} & (16, 100, $\frac{1}{8}$, 0.3) & \textbf{0.46} & (8, 100, $\frac{1}{4}$, 0.3) & \textbf{0.40} & (32, 100, $\frac{1}{16}$, 0.3) & \textbf{0.11} & (64, 100, $\frac{1}{32}$, 0.3) \\
        $L_2$ Distance & 0.28 & (16, 100, $\frac{1}{8}$, 0.3) & 0.39 & (8, 100, $\frac{1}{4}$, 0.3) & 0.29 & (32, 100, $\frac{1}{16}$, 0.3) & 0.06 & (64, 100, $\frac{1}{32}$, 0.3) \\
        Jeffrey Divergence & 0.22 & (16, 10~~, $\frac{1}{8}$, ~10) & 0.33 & (8, 10~~, $\frac{1}{4}$, ~10) & 0.10 & (32, 10~~, $\frac{1}{16}$, ~10) & 0.02 & (64, 10~~, $\frac{1}{32}$, ~10) \\
        Mahalanobis Distance & 0.17 & (16, 10~~, $\frac{1}{8}$, ~10) & 0.30 & (8, 10~~, $\frac{1}{4}$, ~10) & 0.09 & (32, 10~~, $\frac{1}{16}$, ~10) & 0.02 & (64, 10~~, $\frac{1}{32}$, ~10) \\
        \bottomrule
    \end{tabular}
    }
\end{table*}

The choice of similarity or distance metrics in $\cL_{\text{SNN}}$ is a crucial hyperparameter. Tab.\ref{tab:sims-metric} documents the disentanglement performance using different metrics in CLEAR. To make a fair comparison, we select hyperparameter configurations that ensure data reconstruction quality. Cosine similarity achieves the highest gMIG among all the implemented metrics. Interestingly, Jeffrey Divergence and Mahalanobis Distance require significantly different hyperparameter configurations to achieve comparable reconstruction.

\section{Implemented Methods for Mutual Information Minimization}
\label{appx:mim}

\subsection{Variational Upper Bounds}
The MI between $\vz^{(c)}$ and $\vz^{(s)}$ is
\begin{equation}
    I(\vz^{(s)}; \vz^{(c)}) = \E_{p(\vz^{(s)}, \vz^{(c)})} \left[ 
    \log \frac{p(\vz^{(s)}, \vz^{(c)})}{p(\vz^{(s)})p(\vz^{(c)})}
    \right]
\end{equation}
It is intractable. \citet{alemi2016deep} develop a variational upper bound of the MI by introduce an approximated marginal $r(\vz^{(c)})$.
\begin{align}
    I(\vz^{(s)}; \vz^{(c)}) &= \E_{p(\vz^{(s)}, \vz^{(c)})} \left[ 
    \log \frac{p(\vz^{(s)} | \vz^{(c)})}{r(\vz^{(s)})}\right] - \KL\Big(p(\vz^{(s)})\| r(\vz^{(s)})\Big) \notag \\
     &\le \E_{p(\vz^{(s)}, \vz^{(c)})} 
     \left[ \log \frac{p(\vz^{(s)} | \vz^{(c)})}{r(\vz^{(s)})}\right]
\end{align}
Then, we would train a Gaussian probabilistic encoder $q_{\psi}(\vz^{(c)} | \vz^{(s)})$ to approximate $p(\vz^{(c)} | \vz^{(s)})$ through maximum likelihood estimation \citep{alemi2016deep}. Simply assuming isotropic Gaussian will lead to highly biased MI estimates \citep{alemi2016deep, cheng2020club}. Hence, in our work we implemented two alternative methods that avoid choosing inappropriate $r(\cdot)$: L1OutUB, sampled CLUB \cite{poole2019l1out, cheng2020club}.

\begin{align}
    \hat{I}_{\text{vL1Out}} &= \frac{1}{N}\sum_{i=1}^{N}
    \left[ \log \frac{q_{\psi}(\vz_i^{(s)}|\vz_i^{(c)})}{\frac{1}{N-1}\sum_{j\ne i}q_{\psi}(\vz_i^{(s)} | \vz_j^{(c)})} \right] \\
    \hat{I}_{\text{vCLUB-S}} &= \frac{1}{N}\sum_{i=1}^{N} \left[ \log q_{\psi}(\vz_i^{(s)}|\vz_i^{(c)}) - \log q_{\psi}(\vz_{k'_i}^{(s)}|\vz_i^{(c)}) \right], k_i' \sim \text{Unif\{1,...,N\}}
\end{align}

Auxiliary encoder $q_{\psi}(\vz^{(s)} | \vz^{(c)})$ and the main VAE are trained in an alternating fashion, where $q_{\psi}(\vz^{(s)} | \vz^{(c)})$ is typically assigned with a larger learning rate and more steps of gradient updates. 

Interestingly, since MI is symmetric, these methods should be applicable for both $p(\vz^{(s)} | \vz^{(c)})$ and $p(\vz^{(s)} | \vz^{(c)})$ . However, our empirical result shows that only $q_{\psi}(\vz^{(s)} | \vz^{(c)})$ achieves reasonable disentanglement whereas $q_{\psi}(\vz^{(c)} | \vz^{(s)})$ give 0 gMIG. One possible explanation is that $\vz^{(c)}$ exhibits higher informational utility under the regularization of $\cL_{\text{SNN}}^{(c)}$.

\subsection{Total Correlation loss \& Density-Ratio Trick}

We can also minimize the MI using TC loss introduced in \citet{kim2018factorvae, abid2019contrastive}, and \citet{louiset2023sepvae}. Adapt it into our notation

\begin{equation}
    I(\vz^{(s)}; \vz^{(c)}) \approx \frac{1}{N} \sum_{i=1}^{N}
    \mathrm{ReLU}\left( \log \left(
    \frac{D_{\psi}(\vz^{(s)}; \vz^{(c)})}{1 - D_{\psi}(\vz^{(s)}; \vz^{(c)})}
    \right) \right)
\end{equation}

$D_{\psi}$ is an adversarial discriminator that identifies if $(\vz^{(s)}; \vz^{(c)})$ is sampled from joint or the marginals. It calculates the conditional probability $\Pr\Big((\vz^{(s)}; \vz^{(c)}) \sim p(\vz^{(s)}; \vz^{(c)})~ \Big|~\vz^{(s)}, \vz^{(c)} \Big)$ \cite{sugiyama2012density}. In its adversarial learning, the ``real'' samples are the actual latent outputs $(\vz^{(s)}; \vz^{(c)})$ from the VAE encoder while the ``fake'' samples are the shuffled values $(\tilde{\vz}^{(s)}; \tilde{\vz}^{(c)})$. Notably, the shuffling is a permutation along the batch dimension rather than a column swap \citep{louiset2023sepvae}. In practice, this decoupling can be achieved simply by shifting the index of $\vz^{(c)}$ by 1. $D_{\psi}$ and the main VAE are trained in an alternating fashion.

\section{Ablation Study via T-SNE Plots}
\label{appx:ablation}

We perform an ablation study using the Styled-MNIST data to empirically verify the significance of contrastive and anti-contrastive regularization (Appx.~\ref{appx:ablation}). We visualize $\vmu^{(c)}$ and $\vmu^{(s)}$, the variational mean of latent representations, in 2D space using t-SNE \citep{van2008tsne}.

\begin{figure}[h!]
    \centering
    \begin{minipage}{0.45\textwidth}
        \centering
        \begin{minipage}{0.49\textwidth}
            \centering
            \includegraphics[width=\textwidth]{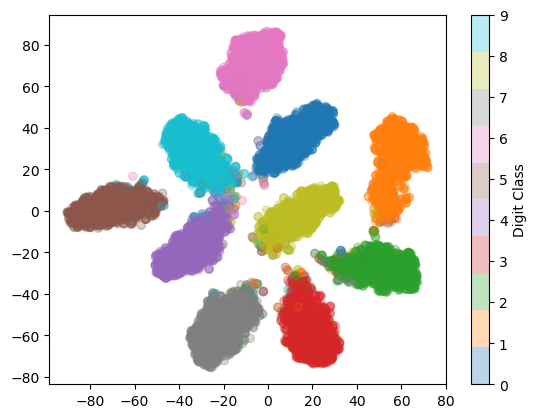}
        \end{minipage}
        \hfill
        \begin{minipage}{0.49\textwidth}
            \centering
            \includegraphics[width=\textwidth]{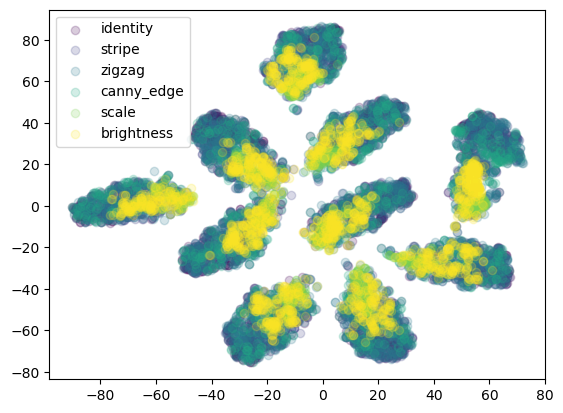}
        \end{minipage}
        \caption*{(a) $\vmu^{(c)}$}
    \end{minipage}
    \hfill
    \begin{minipage}{0.45\textwidth}
        \centering
        \begin{minipage}{0.49\textwidth}
            \centering
            \includegraphics[width=\textwidth]{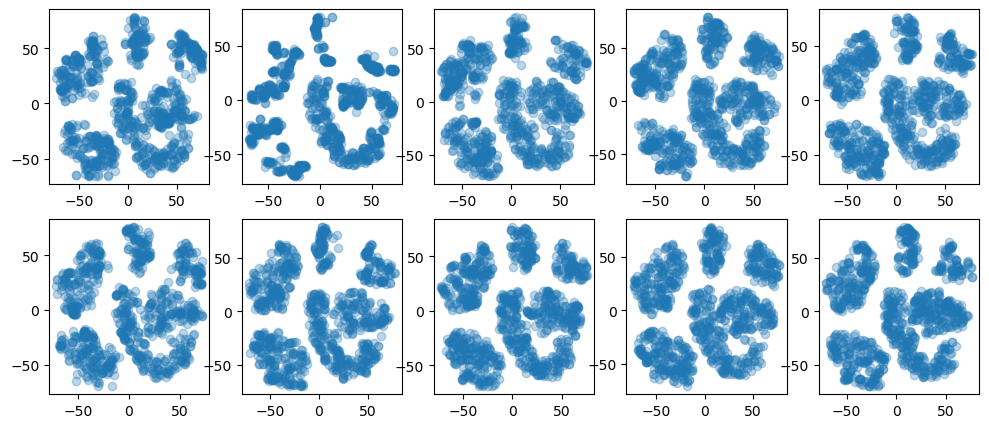}
        \end{minipage}
        \hfill
        \begin{minipage}{0.49\textwidth}
            \centering
            \includegraphics[width=\textwidth]{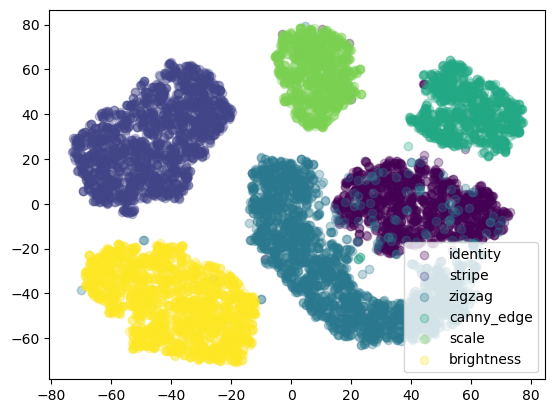}
        \end{minipage}
        \caption*{(b) $\vmu^{(s)}$}
    \end{minipage}
    \caption{$\text{gMIG} = 0.305$. CLEAR-VAE (with $\cL_\mathrm{PS-SNN}^{(c)}$) t-SNE visualization for \textit{content} and \textit{style} latent representations in testing styled-MNIST. The left figure of panel (b) stratifies $\vmu^{(s)}$ according to the digit, with the stratified plots arranged sequentially and in rows. Cosine similarity is used in $\cL_{\text{SNN}}$. The hyperparameter configuration for the training objective function is: $d_{\vz} = 16, \tau = 0.3, \beta = 1/8, \alpha = 100$.}
    \label{fig:tsne-mnist}
\end{figure}

\begin{figure}[h!]
    \centering
    \begin{minipage}{0.45\textwidth}
        \centering
        \begin{minipage}{0.49\textwidth}
            \centering
            \includegraphics[width=\textwidth]{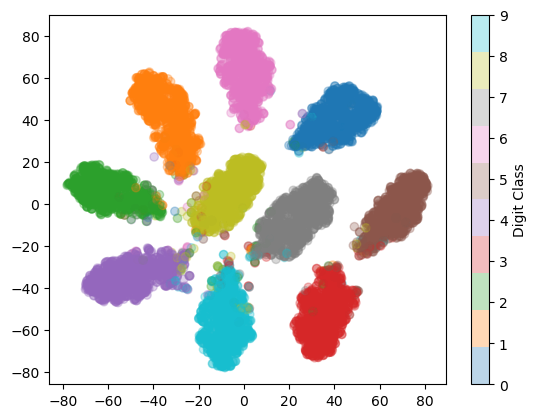}
        \end{minipage}
        \hfill
        \begin{minipage}{0.49\textwidth}
            \centering
            \includegraphics[width=\textwidth]{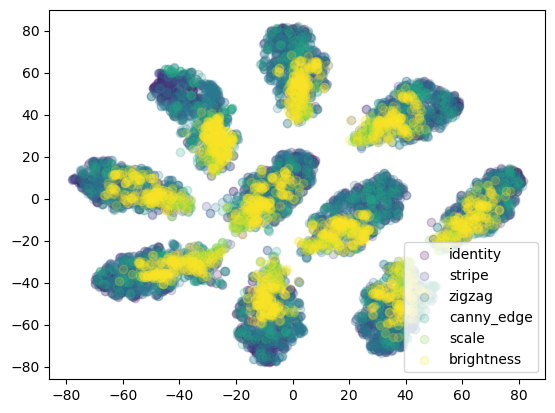}
        \end{minipage}
        \caption*{(a) $\vmu^{(c)}$}
    \end{minipage}
    \hfill    
    \begin{minipage}{0.45\textwidth}
        \centering
        \begin{minipage}{0.49\textwidth}
            \centering
            \includegraphics[width=\textwidth]{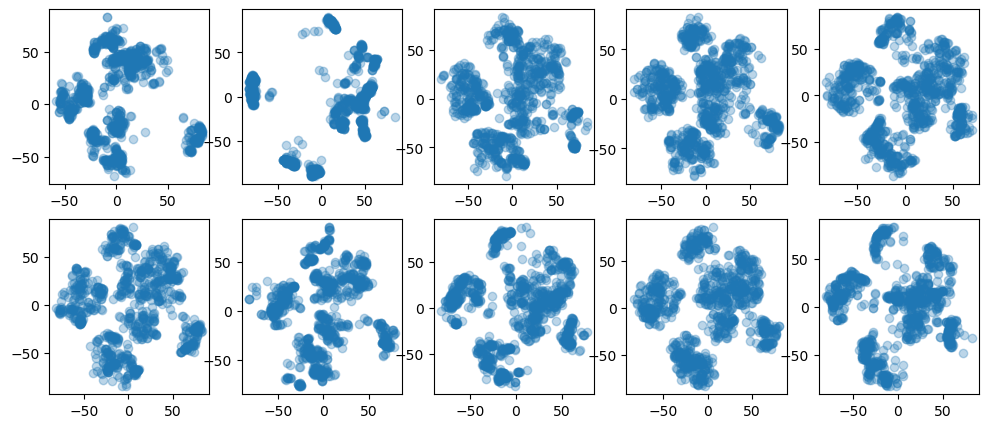}
        \end{minipage}
        \hfill
        \begin{minipage}{0.49\textwidth}
            \centering
            \includegraphics[width=\textwidth]{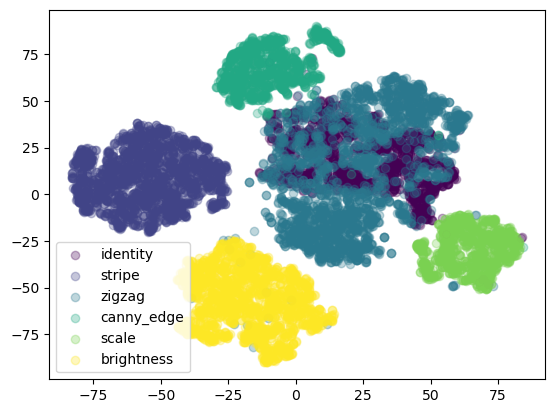}
        \end{minipage}
        \caption*{(b) $\vmu^{(s)}$}
    \end{minipage}
    \caption{$\text{gMIG}=0.265$. CLEAR-VAE (without $\cL_\mathrm{PS-SNN}^{(s)}$) t-SNE visualization for \textit{content} and \textit{style} latent representations in styled-MNIST. The hyperparameter configuration is the same as that of Fig. \ref{fig:tsne-mnist}.}
    \label{fig:tsne-mnist-abl1}
\end{figure}

\begin{figure}[h!]
    \centering
    \begin{minipage}{0.45\textwidth}
        \centering
        \begin{minipage}{0.49\textwidth}
            \centering
            \includegraphics[width=\textwidth]{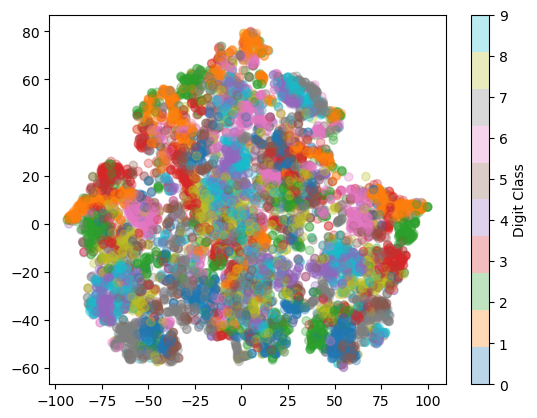}
        \end{minipage}
        \hfill
        \begin{minipage}{0.49\textwidth}
            \centering
            \includegraphics[width=\textwidth]{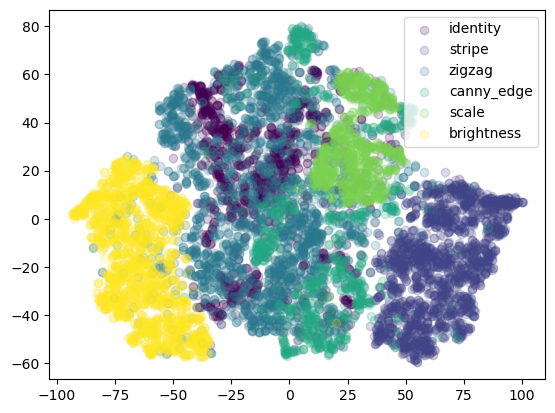}
        \end{minipage}
        \caption*{(a) $\vmu^{(c)}$}
    \end{minipage}
    \hfill
    \begin{minipage}{0.45\textwidth}
        \centering
        \begin{minipage}{0.49\textwidth}
            \centering
            \includegraphics[width=\textwidth]{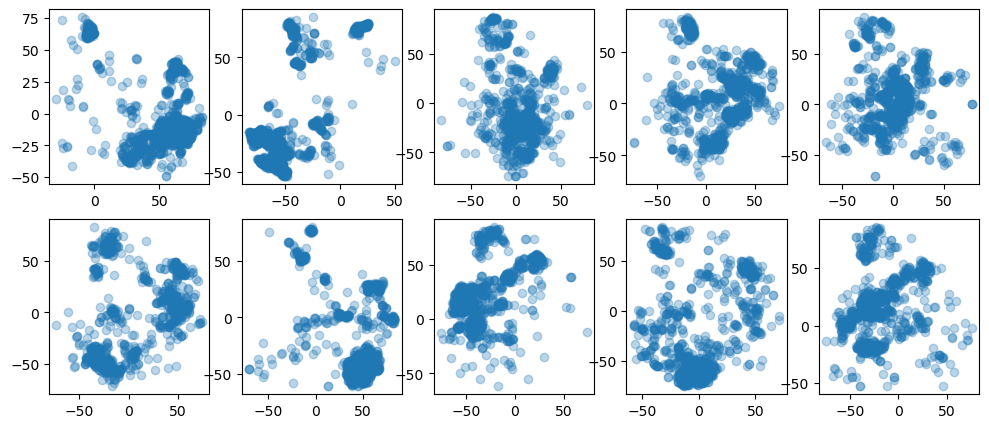}
        \end{minipage}
        \hfill
        \begin{minipage}{0.49\textwidth}
            \centering
            \includegraphics[width=\textwidth]{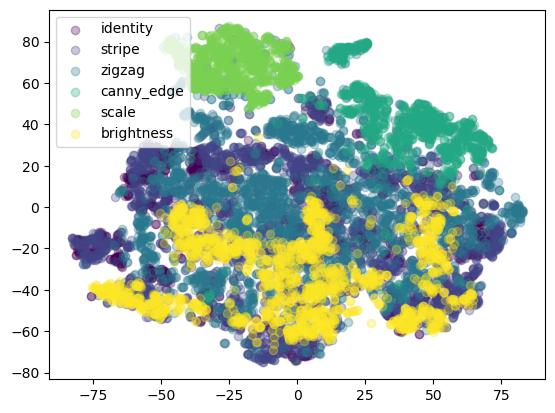}
        \end{minipage}
        \caption*{(b) $\vmu^{(s)}$}
    \end{minipage}
    \caption{$\text{gMIG}=0$. $\beta$-VAE t-SNE visualization for \textit{content} and \textit{style} latent representations in styled-MNIST. he hyperparameter configuration is the same as that of Fig. \ref{fig:tsne-mnist}.}
    \label{fig:tsne-mnist-abl2}
\end{figure}

Here, we use the Styled-MNIST for as an exemplar illustration. Fig. \ref{fig:tsne-mnist} visualizes clear clustering patterns in the 2D t-SNE space. Panel (a) shows the clustering of $\vmu^{(c)}$ colored by \textit{content} and \textit{style}, respectively. Panel (b) shows the $\vmu^{(s)}$ distributions stratified by \textit{content} and the clustering of $\vmu^{(s)}$ colored by \textit{style}. Without the contrastive and the anti-contrastive regularization, the representations of \textit{content} and \textit{style} are entangled in the latent space. When we remove both $\cL_{\text{SNN}}^{(c)}$ and $\cL_{\text{PS-SNN}}^{(s)}$, a CLEAR-VAE will be reduced to a $\beta$-VAE. Fig. \ref{fig:tsne-mnist-abl1} indicates that the model fails to disentangle \textit{content} and \textit{style}. If we remove $\cL_{\text{PS-SNN}}^{(s)}$, the CLEAR-VAE is still able to learn the representation of content and disentangle it from style to some extent, but Fig. \ref{fig:tsne-mnist-abl2} shows that it is incapable of extracting details (e.g. zig-zag lines) in style features. 

\newpage
\section{Implementation Details}
\subsection{Computing Resource}
The experiments were conducted on Google Colab NVIDIA L4 GPUs. The experiments were also tested on NVIDIA GeForce RTX 4080 Laptop GPU for reproducibility. A mid-range laptops with 8GB to 12GB of VRAM can carry out all the experiments.

\subsection{Model architecture}
\label{appx:arch}

Encoder modules (including all VAE encoders and CNN feature extraction head) have a common structure: sequential blocks of \texttt{Conv2D-BN2D-ReLU}. Decoder modules have the same structure: sequential blocks of \texttt{ConvTranspose2D-BN2D-ReLU}. The parameters are initialized using He initialization \citep{he2015delving} to stabilize training and accelerate convergence for Camelyon17-WILD dataset, as we observe extremely volatile change in gMIG during train. The hyperparameters used in \texttt{Conv2D} and \texttt{ConvTranspose2D} depend on the shape of the modality, which are listed as follows.

\begin{enumerate}
    \item $(\text{C}\times28\times28)$: \begin{enumerate}
        \item \texttt{Conv2D}: 
        
        [C, 32, 3, 2, 1], [32, 64, 3, 2, 1], [64, 128, 3, 2, 1]
        \item \texttt{ConvTranspose2D}:
        
        [128, 64, 3, 2, 1], [64, 32, 3, 2, 1, 1], [32, C, 3, 2, 1, 1]
    \end{enumerate}
    \item $(\text{C}\times64\times64)$: \begin{enumerate}
        \item \texttt{Conv2D}: 
        
        [C, 32, 4, 2, 1], [32, 64, 4, 2, 1], [64, 128, 4, 2, 1], [128, 256, 4, 2, 1], [256, 512, 4, 2, 1]
        \item \texttt{ConvTranspose2D}:
        
        [512, 256, 4, 2, 1], [256, 128, 4, 2, 1], [128, 64, 4, 2, 1], [64, 32, 4, 2, 1], [32, C, 4, 2, 1]
    \end{enumerate}
\end{enumerate}

In CLEAR-L1OutUB and CLEAR-CLUB-S, the auxiliary network are made of 2-layer MLPs for $\vmu$ and $\Sigma$, receptively. The input dimension should match $\vz^{(c)}$'s dimension.
In CLEAR-TC, the TC discriminator is a 2-layer MLP classifier. The input dimension should match dimension of the entire latent feature vector.

For classification tasks, all methods employ 2-layer MLP classification heads with identical shapes. The input dimension should match the $z^{(c)}$'s dimension

\newpage

\section{Model comparison through Swapping experiments}
\label{appx:more-figs}

\begin{figure}[h!]
    \centering
    \begin{minipage}{0.3\textwidth}
        \centering
        \includegraphics[width=\linewidth]{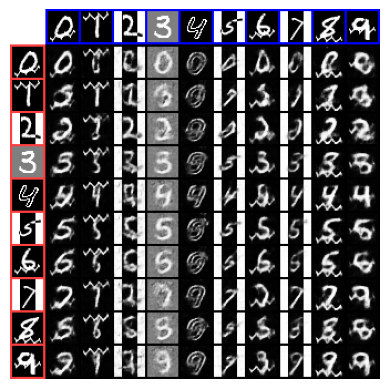}
        \caption*{(a) GVAE}
    \end{minipage}
    \begin{minipage}{0.3\textwidth}
        \centering
        \includegraphics[width=\linewidth]{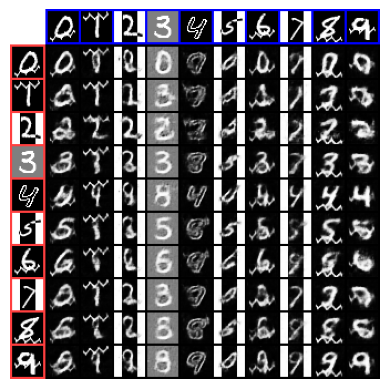}
        \caption*{(b) ML-VAE}
    \end{minipage}
    \begin{minipage}{0.3\textwidth}
        \centering
        \includegraphics[width=\linewidth]{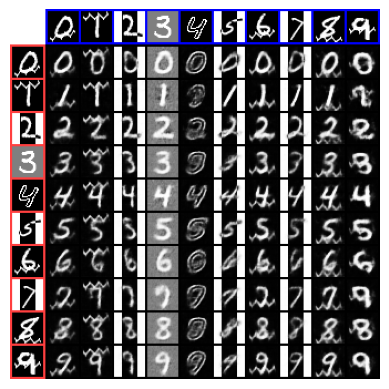}
        \caption*{(c) CLEAR-TC}
    \end{minipage}
    
    \vspace{0.5cm}  % Space between rows
    
    \begin{minipage}{0.3\textwidth}
        \centering
        \includegraphics[width=\linewidth]{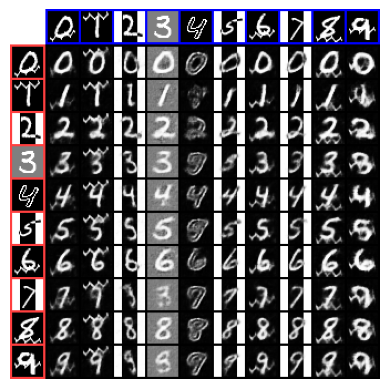}
        \caption*{(d) CLEAR-PS (\textbf{ours})}
    \end{minipage}
    \begin{minipage}{0.3\textwidth}
        \centering
        \includegraphics[width=\linewidth]{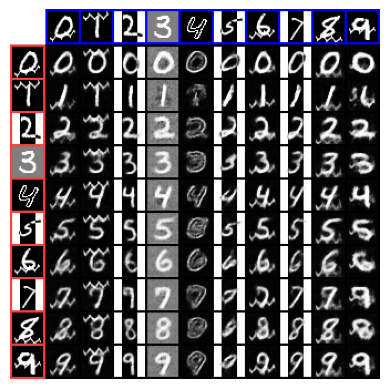}
        \caption*{(e) CLEAR-L1OutUB}
    \end{minipage}
    \begin{minipage}{0.3\textwidth}
        \centering
        \includegraphics[width=\linewidth]{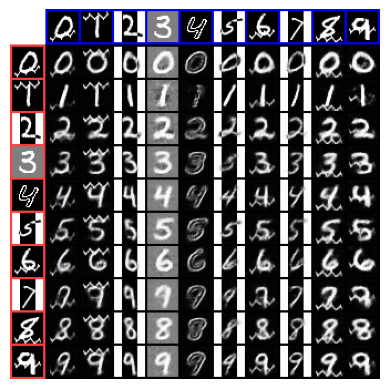}
        \caption*{(f) CLEAR-CLUB-S}
    \end{minipage}
    
    \caption{Comparing swapping experiments using Styled-MNIST}
    \label{fig:appx-styled-mnist}
\end{figure}

\begin{figure}[h!]
    \centering
    \begin{minipage}{0.3\textwidth}
        \centering
        \includegraphics[width=\linewidth]{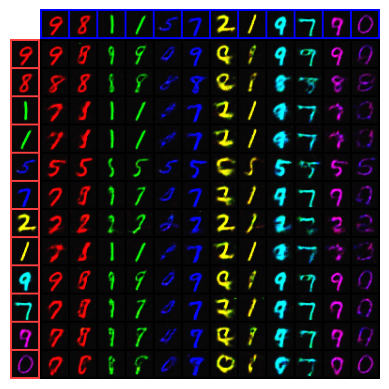}
        \caption*{(a) GVAE}
    \end{minipage}
    \begin{minipage}{0.3\textwidth}
        \centering
        \includegraphics[width=\linewidth]{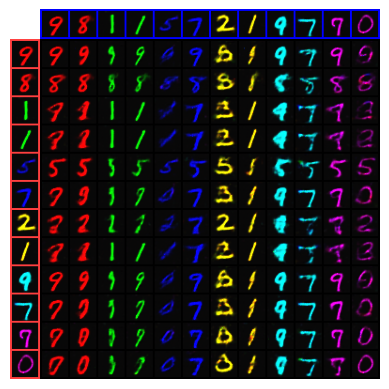}
        \caption*{(b) ML-VAE}
    \end{minipage}
    \begin{minipage}{0.3\textwidth}
        \centering
        \includegraphics[width=\linewidth]{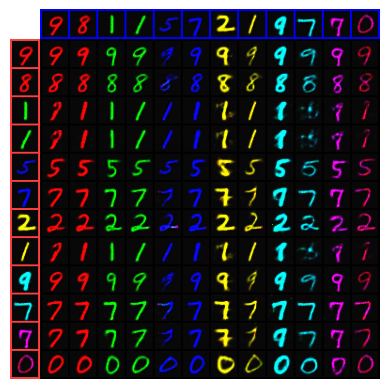}
        \caption*{(c) CLEAR-TC}
    \end{minipage}
    
    \vspace{0.5cm}  % Space between rows
    
    \begin{minipage}{0.3\textwidth}
        \centering
        \includegraphics[width=\linewidth]{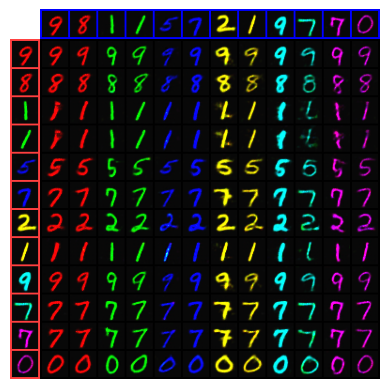}
        \caption*{(d) CLEAR-PS (\textbf{ours})}
    \end{minipage}
    \begin{minipage}{0.3\textwidth}
        \centering
        \includegraphics[width=\linewidth]{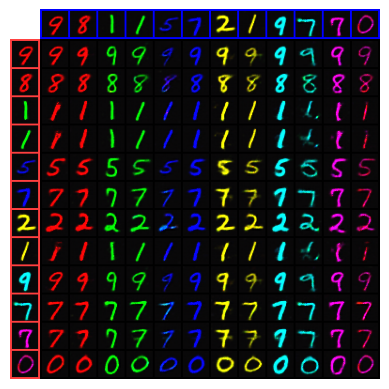}
        \caption*{(e) CLEAR-L1OutUB}
    \end{minipage}
    \begin{minipage}{0.3\textwidth}
        \centering
        \includegraphics[width=\linewidth]{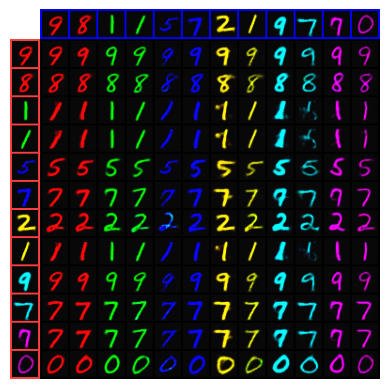}
        \caption*{(f) CLEAR-CLUB-S}
    \end{minipage}
    
    \caption{Comparing swapping experiments using Colored-MNIST}
    \label{fig:appx-colored-mnist}
\end{figure}

\begin{figure}[h!]
    \centering
    \begin{minipage}{0.3\textwidth}
        \centering
        \includegraphics[width=\linewidth]{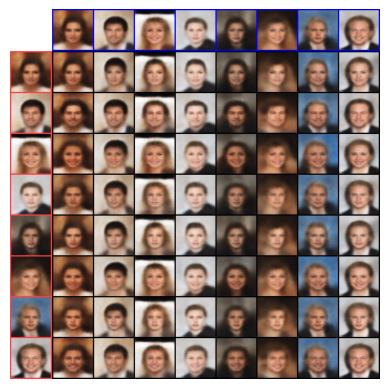}
        \caption*{(a) GVAE}
    \end{minipage}
    \begin{minipage}{0.3\textwidth}
        \centering
        \includegraphics[width=\linewidth]{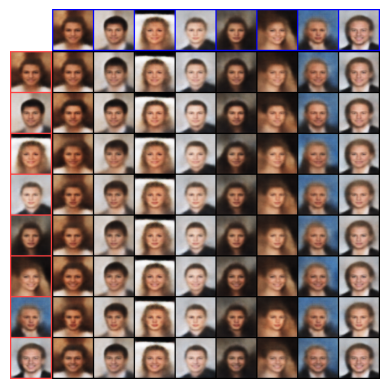}
        \caption*{(b) ML-VAE}
    \end{minipage}
    \begin{minipage}{0.3\textwidth}
        \centering
        \includegraphics[width=\linewidth]{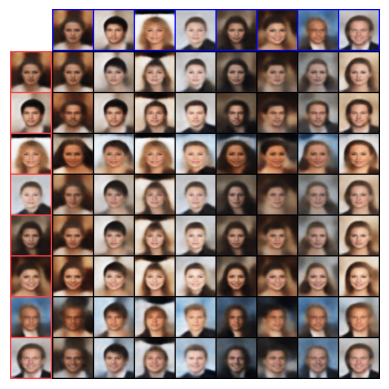}
        \caption*{(c) CLEAR-TC}
    \end{minipage}
    
    \vspace{0.5cm}  % Space between rows
    
    \begin{minipage}{0.3\textwidth}
        \centering
        \includegraphics[width=\linewidth]{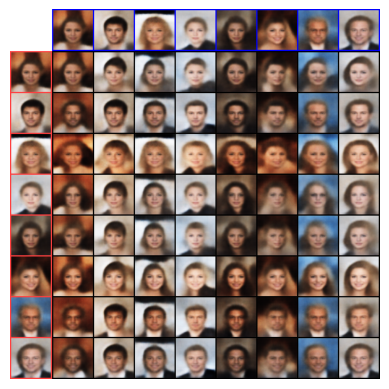}
        \caption*{(d) CLEAR-PS (\textbf{ours})}
    \end{minipage}
    \begin{minipage}{0.3\textwidth}
        \centering
        \includegraphics[width=\linewidth]{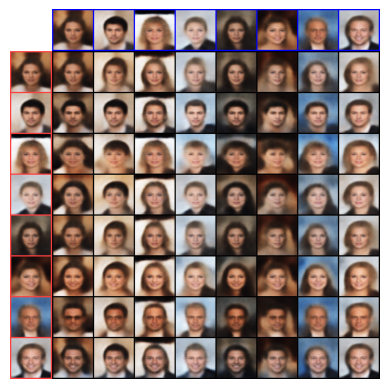}
        \caption*{(e) CLEAR-L1OutUB}
    \end{minipage}
    \begin{minipage}{0.3\textwidth}
        \centering
        \includegraphics[width=\linewidth]{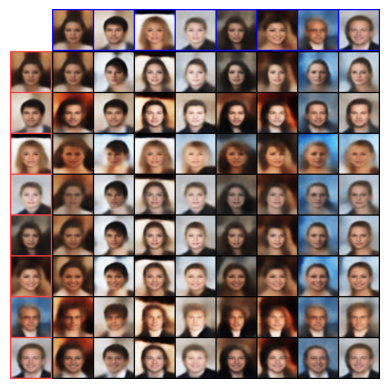}
        \caption*{(f) CLEAR-CLUB-S}
    \end{minipage}
    
    \caption{Comparing swapping experiments using CelebA}
    \label{fig:main}
\end{figure}

\end{document}